\PassOptionsToPackage{table}{xcolor}
\PassOptionsToPackage{activate={true,nocompatibility}}{microtype}

\documentclass[10pt,letterpaper]{wlscirep}

\usepackage{array}
\usepackage{booktabs}
\usepackage{multirow}
\usepackage{float}
\usepackage{changepage}
\usepackage{needspace}
\usepackage{placeins}

\newcolumntype{+}{!{\vrule width 2pt}}

\newlength\savedwidth

\hypersetup{colorlinks=true, allcolors=blue}

\title{Limited Structural Reliability in Public Educational Prediction Benchmarks:\\A Four-Dimension Audit of Seven Datasets}

\author[1,2]{Yan Ma}
\author[3,*]{Lizhuo Zhang}
\affil[1]{School of Foreign Studies, Changsha University of Science and Technology, Changsha 410114, China}
\affil[2]{School of Education, Hunan Agricultural University, Changsha 410128, China}
\affil[3]{School of Information and Intelligence, Hunan Agricultural University, Changsha 410128, China}
\affil[*]{zhanglizhuo@hunau.edu.cn}


\begin{document}
\flushbottom

\begin{abstract}
Across seven public educational prediction datasets, three passed all four
pre-modeling reliability checks; the remaining four either failed group-aware
generalization tests or lacked the provenance metadata needed to run them. One
dataset was initially classified as failing but corrected after excluding
group-identifier features from the holdout matrix, demonstrating that the audit
can distinguish genuine cross-group confounding from feature-encoding artifacts.
Each dataset was audited before model optimization using four checks: baseline gap,
split instability, null separation, and metadata adequacy under group-aware
holdout. The dominant failure mode was not weak iid performance alone but
cross-group fragility: in the clearest case, UCI Student declined from iid
$R^2 = 0.242$ to group-holdout $R^2 = -0.097$, while Higher Ed collapsed from $0.041$ to $-8.79$. Increasing model complexity did
not remove this pattern: ensemble models improved structurally sound datasets
but amplified instability or failed under group holdout on fragile ones. An
exploratory cross-dataset comparison further showed that stronger profiles
clustered in larger, richer-grouped, performance-proximal datasets, while
random-split performance severely overstated deployable signal in fragile
datasets. Classification-metric sensitivity analyses reached the same substantive
conclusions. The results show that benchmark reliability in educational AI is
constrained less by algorithm choice than by data structure, group heterogeneity,
and evaluation design. A reusable pre-modeling audit offers a minimum quality gate before public
educational datasets support strong benchmark or deployment claims.

\end{abstract}
\maketitle
\section{Introduction}

Educational AI increasingly uses predictive models to inform teaching evaluation, student intervention, and institutional policy. Yet many reported gains remain hard to interpret because benchmark reliability is rarely examined before modeling begins. In computer vision and NLP, benchmark-audit studies have become routine \citep{Northcutt2021,Recht2019,Bouthillier2021}; in educational prediction, comparable stress tests remain uncommon. Educational datasets are often smaller and more nested than the vision benchmarks where much audit work has been conducted, making them more vulnerable to group-level confounding.

This gap is consequential. Educational datasets typically contain nested group structure---students within classrooms, classrooms within schools, schools within districts---that creates confounding risks absent from many standard ML benchmarks. If teacher identity, school culture, cohort composition, course presentation, or assessment pipeline is under-specified, a model can learn contextual shortcuts rather than generalizable pedagogical signal. Under those conditions, high correlation is not evidence of instructional quality; it may reflect unresolved group-level confounding, label noise, non-independent partitions, or distribution shift across educational contexts.

The growing use of LLM-derived scoring in educational assessment introduces additional risk \citep{Zheng2023,Wang2024,Huang2024llm}. When both features and targets originate from the same automated pipeline, apparent predictive performance may reflect shared artifacts rather than pedagogical signal.

Prior work in two areas motivates this audit. In machine-learning benchmarking, label errors \citep{Northcutt2021}, distribution shift \citep{Recht2019}, and split-variance \citep{Bouthillier2021} have all been shown to destabilize benchmark conclusions; yet nearly all such audit work targets computer-vision or NLP benchmarks, and comparable stress tests of educational prediction datasets remain scarce. In educational data mining and learning analytics, many studies predict student outcomes \citep{Romero2020,Baker2014,Gardner2019}, and recent work has emphasized cross-institutional generalization \citep{Chejara2023,Chejara2024}; in practice, however, iid evaluation without explicit group-aware holdout remains common.

Two further traditions clarify why this matters. Educational measurement theory asks whether scores generalize across facets and contexts: Generalizability Theory \citep{Brennan2001} treats train-test partitioning as a variance facet, and the validity-argument framework \citep{Messick1995,Kane2006} requires generalizability evidence before score interpretations are warranted. The audit translates these concerns into computational checks by treating group membership as a potential validity threat. Data-documentation work---Datasheets for Datasets \citep{Gebru2021}, data cards, and data-cascade analyses \citep{Sambasivan2021}---shows how upstream quality failures propagate through ML pipelines.

The question is straightforward: are public educational prediction benchmarks structurally reliable enough to support generalizable modeling claims? To answer it, seven public datasets are audited with a common four-dimension framework covering baseline gap (does the model beat a trivial reference?), split instability (do conclusions change across random partitions?), null separation (does observed performance separate from permuted labels?), and metadata adequacy (does performance survive group-aware holdout when the requisite grouping variables are available?). The protocol functions as a pre-modeling quality gate: before model comparisons are reported, a dataset can be screened for stable, non-null, and group-robust signal. In the terminology of meta-science \citep{Ioannidis2005}, the paper asks not only ``how well can models predict?'' but also ``when do benchmark claims remain credible under stronger tests of reliability and generalizability?''

Three research questions guide the analysis. \textbf{RQ1:} How often do public educational prediction datasets satisfy all four audit dimensions under a common evaluation framework? \textbf{RQ2:} When datasets fail, which failure modes dominate: weak baseline advantage, split sensitivity, lack of null separation, or cross-group fragility? \textbf{RQ3:} Do these conclusions persist when model complexity and task-appropriate classification metrics are varied? Across the seven audited datasets, three pass all four checks, indicating that the main threat to benchmark validity in educational prediction lies in data structure and evaluation design rather than in insufficient model sophistication. The paper contributes an empirical finding about cross-group fragility in current educational benchmarks and a reusable audit procedure for screening future datasets before strong benchmark claims are made.

\section{Materials and Methods}

\subsection{Audit Protocol}
\begin{figure}[!t]
\centering
\includegraphics[width=0.95\textwidth]{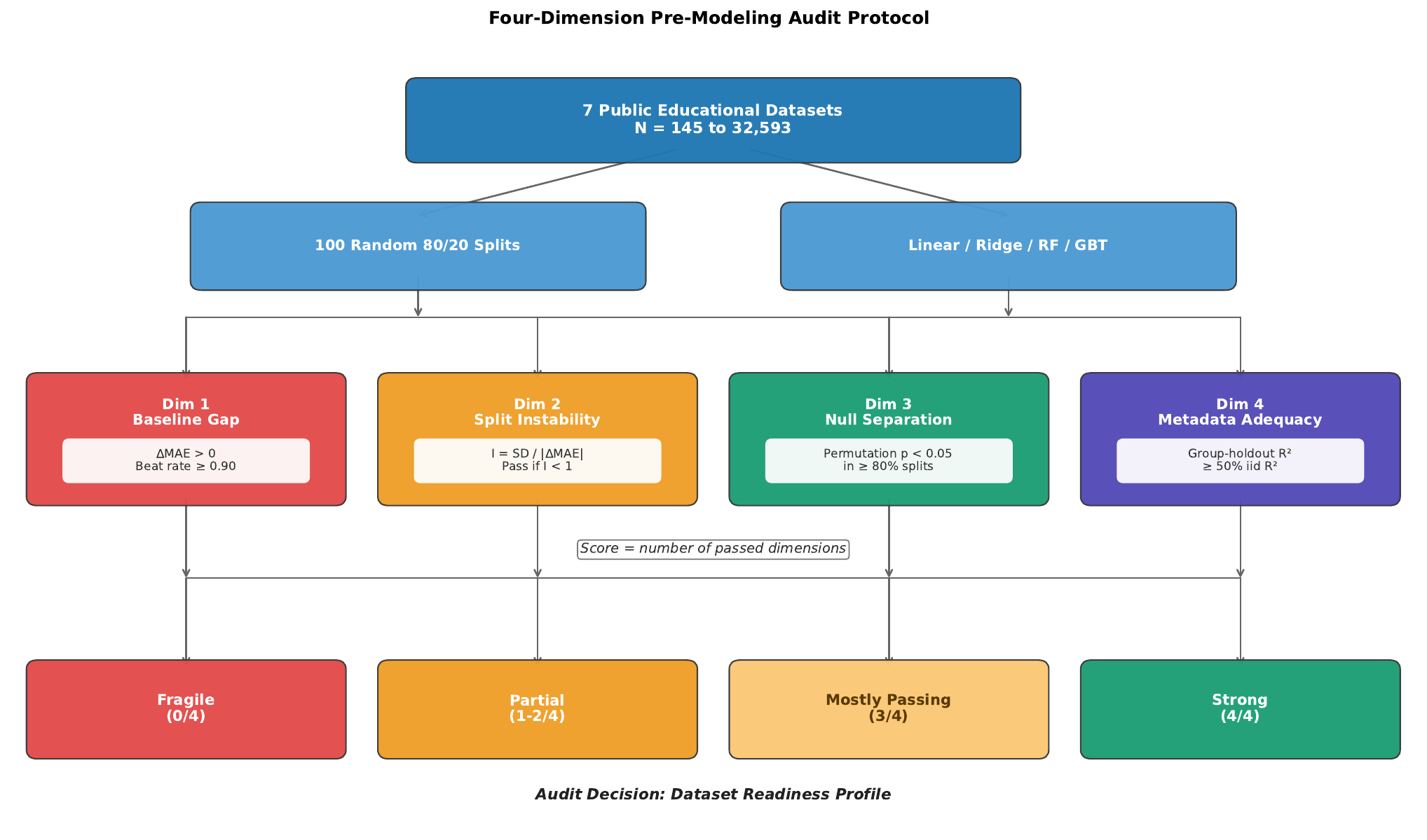}
\caption{Four-dimension pre-modeling audit protocol. Seven public educational datasets are evaluated by 100 random 80/20 splits across four model families (linear, ridge, RF, GBT). Each dataset is then scored on baseline gap, split instability ($I = \mathrm{SD}/|\Delta_{\mathrm{MAE}}|$), null separation (permutation $p < 0.05$ across $\geq 80\%$ of splits), and metadata adequacy (group-holdout retention $\geq 50\%$). The number of passed dimensions defines the readiness profile (Fragile / Partial / Mostly Passing / Strong).}
\label{fig:protocol}
\end{figure}

The audit examines each dataset along four diagnostic dimensions using parsimonious linear baselines and standard statistical diagnostics. (1) \textbf{Baseline gap} measures whether the main predictive models improve on a trivial reference such as mean-only prediction. (2) \textbf{Split instability} quantifies sensitivity to admissible train-test partitions, following the concern raised by Bouthillier et al.~\citep{Bouthillier2021} that random-seed and split variation can inflate apparent method differences. (3) \textbf{Null-separation status} records whether observed association statistics separate from an empirical permutation null, analogous to the permutation-based checks used in benchmark-reliability studies \citep{Northcutt2021}. (4) \textbf{Metadata adequacy} records whether grouping information (e.g., teacher, school, or cohort identifiers) needed for leakage-resistant validation is actually available in the release and, when available, whether performance survives group-aware holdout. Dataset readiness is interpreted conservatively: a release is not considered ready for strong benchmark claims when baseline advantage is trivial or unstable, null separation is absent or inconsistent, or metadata adequacy fails. The four dimensions are not aggregated into a single scalar score because the appropriate weighting would depend on domain context. Based on the number of dimensions passed, the audit assigns each dataset a structured profile: \textbf{Fragile} (0/4), \textbf{Partial} (1--2/4), \textbf{Mostly Passing} (3/4), or \textbf{Strong} (4/4). Fig~\ref{fig:protocol} summarizes the protocol workflow from raw datasets through the four diagnostic dimensions to the final readiness profile.

\subsection{Operational Decision Rules}
The analysis adopts operational thresholds rather than universal normative cutoffs. Let $\Delta_{\mathrm{MAE}} = \mathrm{MAE}_{\text{mean}} - \mathrm{MAE}_{\text{model}}$ (positive = improvement over trivial baseline) and the split-instability ratio $I = \mathrm{SD}(\mathrm{MAE}_{\text{model}}) / (|\Delta_{\mathrm{MAE}}| + \varepsilon)$, with $\varepsilon = 10^{-8}$. Baseline gap is supported when $\Delta_{\mathrm{MAE}} > 0$ with beat rate $\geq 0.90$ across $\geq 50$ splits. Split instability fails when $I \geq 1$, that is, when between-split variability equals or exceeds the mean advantage. Null separation is supported when permutation $p < 0.05$ is consistent across $\geq 80\%$ of tested splits. Metadata adequacy passes when group-aware holdout retains $\geq 50\%$ of iid $R^2$; it fails when $R^2$ collapses under group holdout or when grouping variables are unavailable. For the primary pass/fail determination of metadata adequacy, the unregularized linear-model group-holdout $R^2$ is used as the common reference because the same linear baseline also anchors $I$, $\Delta_{\mathrm{MAE}}$, beat rate, and permutation testing.

The rationale for each threshold is empirical rather than universal. The $I = 1$ cutoff has a direct interpretation: when between-split variability equals the reported gain, a single random split has roughly coin-flip odds of showing improvement. The beat-rate threshold of 0.90 follows the same logic: if a model fails to outperform a trivial baseline in more than 10\% of splits, the gain is not reliably reproducible. The permutation-consistency threshold of 80\% mirrors the conventional power standard ($1 - \beta = 0.80$). The 50\% group-holdout retention threshold marks the point below which more than half of the iid signal is attributable to within-group structure. These thresholds are empirically motivated starting points rather than universal normative cutoffs. Supplementary Figure~S3 shows that the main conclusions are robust to threshold variation within $\pm 20\%$ of each adopted value, confirming that the reported profiles are not artifacts of borderline cutoff choices. The thresholds are summarized in Table~\ref{tab:decision_criteria}.

\subsection{Datasets}
Seven public educational datasets are audited (Table~\ref{tab:dataset_overview}). They were selected to span sample sizes from $N = 145$ to $32{,}593$, multiple educational settings (K-12, higher education, entrance examination, and online learning), continuous and discrete targets, and grouping structures ranging from 2 to 22 groups, plus one release with no grouping metadata. The aim is not exhaustive coverage of educational AI datasets, but a stress test of the audit protocol across the dataset types most commonly used in educational prediction studies.

\begin{table}[H]
\centering
\caption{Overview of the seven audited educational datasets.}
\label{tab:dataset_overview}
\resizebox{\textwidth}{!}{%
\begin{tabular}{l r r l l l l}
\toprule
Dataset & $N$ & Features & Target & Type & Grouping variable & Groups \\
\midrule
MM-TBA & 186 & 13 & Lecture quality (0--7.5) & Continuous & None available & --- \\
Higher Ed & 145 & 31 & Course grade (0--7) & Continuous & Course ID & 9 \\
xAPI-Edu & 480 & 72 & Engagement (L/M/H $\to$ 0/1/2) & Ordinal 3-class & Topic & 12 \\
Entrance Exam & 666 & 49 & Performance (4-class $\to$ 0--3) & Ordinal 4-class & Education board & 3 \\
UCI Student & 649 & 56 & Final grade $G3$ (0--20) & Continuous & School & 2 \\
Dropout & 3,630 & 36 & Graduate/Dropout (1/0) & Binary & Course & 17 \\
OULAD & 32,593 & 44 & Pass-or-Distinction vs. Fail-or-Withdrawn (1/0) & Binary & Course-presentation & 22 \\
\bottomrule
\end{tabular}}%
\end{table}

\textbf{(1) MM-TBA} \citep{Huang2025}: 186 lecture segments, 13 features (8 transcript-derived, 5 metadata covariates), target is the mean of GPT-generated rubric subscores, and no teacher-identity metadata are available. \textbf{(2) Higher Ed} (Yilmaz and Sekeroglu~\citep{Yilmaz2020}; UCI Machine Learning Repository ID 856): $N=145$, 31 features, continuous course grade (0--7), grouped by course (9 courses). \textbf{(3) xAPI-Edu-Data} \citep{Amrieh2016}: $N=480$, 72 features after one-hot encoding, three-class engagement target mapped to 0/1/2, grouped by topic (12 subjects). \textbf{(4) Entrance Exam} (Bora and Dey~\citep{Bora2021}; UCI Machine Learning Repository ID 582): $N=666$, 49 features after one-hot encoding, four-class target mapped to 0--3, grouped by education board (3 boards). \textbf{(5) UCI Student} \citep{Cortez2008}: Portuguese-language subset, $N=649$, 56 features after one-hot encoding, continuous final grade $G3 \in [0, 20]$ (with $G1$ and $G2$ excluded), grouped by school (GP: $n=423$; MS: $n=226$). \textbf{(6) Dropout} (Realinho et al.~\citep{Realinho2022}; UCI ID 697): $N=3{,}630$ after excluding ``Enrolled'' students, 36 features, binary Graduate/Dropout target, grouped by course (17 programs). \textbf{(7) OULAD} \citep{Kuzilek2017}: $N=32{,}593$, 44 features after one-hot encoding, binarized pass-or-distinction vs. fail-or-withdrawn target, grouped by course-presentation cohort (22 cohorts).

Table~\ref{tab:dataset_overview} reports post-encoding feature counts (Supplementary Table~S2 reports additional exploratory data characteristics). All categorical variables are one-hot encoded. Within each split, all features are Z-score standardized using parameters fitted exclusively on the training partition and applied to the corresponding test partition. Classification targets are mapped to equally spaced integers so that MAE, $R^2$, and Pearson $r$ can be computed consistently across all seven datasets. This regression framing is used for cross-task audit comparability, not as a psychometric claim that ordinal class intervals are intrinsically equal.

\subsection{Models and Evaluation}
\label{sec:models_eval}
All seven datasets are evaluated under a common model stack: mean-only baseline, unregularized linear regression, ridge regression, Random Forest, and Gradient Boosting. The core audit reports the linear model as the reference because it is the least forgiving detector of weak signal and group-sensitive leakage, while ridge and ensemble models are retained to determine whether failures are structurally invariant or partly remediable by regularization or nonlinear fitting.

The linear model is used as the primary reference for the metadata-adequacy judgment because it is the least forgiving detector of group-sensitive leakage: an unregularized linear fit maximizes the contrast between within-group and cross-group signal. Regularized or ensemble models that recover under group holdout (e.g., ridge and ensemble models on Entrance Exam) indicate that the metadata-adequacy failure is partly remediable through modeling choices, but the audit protocol is designed to flag the structural risk inherent in the benchmark rather than to certify whether some model in the hyperparameter space can compensate for it. Using the best-performing model for the pass/fail judgment would create a perverse incentive: datasets with missing grouping metadata could be claimed as ``passing'' simply by deploying a model complex enough to overfit group-specific noise.

Evaluating all seven datasets under a common regression framing ensures cross-dataset comparability of the audit metrics (MAE, $R^2$, Pearson $r$). For binary and ordinal targets, outcomes are mapped to equally spaced integers, and a dedicated sensitivity analysis (Section~\ref{sec:class_sensitivity}) verifies that the audit profiles are robust to classification-appropriate metrics. Integer mapping imposes an equal-interval assumption on ordinal scales; this is a pragmatic choice for cross-task comparability rather than a psychometric claim.

The following leakage-prevention measures were applied uniformly: (i) all standardization (Z-score) parameters were fitted exclusively on the training partition of each split and applied to the corresponding test partition; (ii) one-hot encoding was fitted on the full dataset but applies only to categorical variables with fixed levels, introducing no lookahead from the test set; (iii) for each dataset, the absence of duplicate or near-duplicate rows was verified (none found); (iv) the UCI Student adapter excludes $G1$ and $G2$ (intermediate-term assessments that post-date $G3$) to prevent target-proxy leakage; (v) for OULAD, assessment scores are treated as features, and all rows in a given split were verified to be temporally coherent within the same course-presentation cohort.

The evaluation procedure has three stages. \textbf{(1) Repeated random splits}: each dataset is evaluated on 100 random 80/20 train-test splits. For every split, models are trained on the training partition and evaluated on the held-out partition. The main summary statistics are MAE, $R^2$, Pearson $r$, beat rate against the mean baseline, and the instability ratio $I$. \textbf{(2) Permutation null}: on 30 of the 100 splits, a 500-draw permutation null is estimated by shuffling labels and refitting the linear model. The rate at which the observed linear $R^2$ exceeds the 95th percentile of the null distribution is recorded as the permutation significance rate. This produces 15{,}000 null fits per dataset and reduces the binomial uncertainty of the null-separation dimension while remaining computationally tractable. \textbf{(3) Group-aware holdout}: where grouping metadata are available, leave-one-group-out holdout is performed for all groups with sufficient size ($n \geq 10$). The model is trained on all remaining groups and tested on the held-out group. Mean and worst-case group-holdout $R^2$ are recorded across all four non-trivial model types. For datasets in which group membership was encoded as one-hot features in the original feature matrix (UCI Student: \texttt{school\_GP}, \texttt{school\_MS}; xAPI-Edu: \texttt{Topic\_*}; Entrance Exam: \texttt{Class\_ten\_education\_*}), group-identifier columns were excluded from the feature matrix prior to leave-one-group-out training to avoid zero-variance predictors in the training partition. For datasets in which group membership was encoded as one-hot features (UCI Student, xAPI-Edu, Entrance Exam), IID resampling also uses the group-identifier-free feature matrix to ensure a fair apples-to-apples comparison with group-holdout results; for all other datasets, the full feature set is retained for IID resampling. For MM-TBA, where no grouping variable is available, this dimension is recorded as failed by default. The grouping variable for each dataset is documented in Table~\ref{tab:dataset_overview}.

This correction primarily affected xAPI-Edu, whose apparent linear-model collapse (previously reported as group-holdout $R^2 = -2.03$) was entirely attributable to zero-variance topic indicators in the training partition. After correction, xAPI-Edu achieves linear group-holdout $R^2 = 0.484$, sufficient to pass the metadata-adequacy threshold, and is reclassified from Mostly Passing (3/4) to Strong (4/4). This reclassification is itself informative: it demonstrates that the audit framework can distinguish genuine cross-group fragility from numerical artifacts introduced by group-collinear feature encoding.

\subsection{Classification-Metric Sensitivity Check}
\label{sec:class_sensitivity}
Because four of the seven datasets have binary or ordinal targets (OULAD, Dropout, xAPI-Edu, Entrance Exam), a sensitivity analysis re-evaluates those datasets under classification-appropriate metrics using Logistic Regression, Random Forest, and Gradient Boosting classifiers. Accuracy and macro-averaged AUC-ROC are reported under both iid resampling and group-aware holdout. An instability ratio $I_{\mathrm{acc}}$ is computed analogously to $I$ but using accuracy in place of MAE. The goal is to test whether the audit conclusions derived from the regression framing are robust to metric choice.

All audit computations are CPU-scale and complete within minutes per dataset.

\section{Results}

Unless otherwise noted, all $\pm$ values denote cross-split standard deviations across 100 repeated 80/20 splits. The audit was applied uniformly to all seven datasets using five models (mean baseline, linear regression, ridge regression, Random Forest, Gradient Boosting) with identical resampling, permutation-null, and group-holdout procedures.

\subsection{Cross-Dataset Audit Summary}

Table~\ref{tab:seven_dataset_audit} presents the main audit results across all seven datasets, ordered by instability ratio.

\begin{table*}[htbp]
\centering
\caption{Seven-dataset audit comparison. $I_{\mathrm{lin}}$ and $I_{\mathrm{gbt}}$ are instability ratios for linear and Gradient Boosting models. Beat rate is the fraction of 100 splits where the linear model outperforms the mean baseline. Perm\% is the fraction of tested splits with permutation $p < 0.05$. Group $R^2$ is the mean leave-one-group-out $R^2$ for the linear model; iid $R^2$ is the mean $R^2$ under random splitting.}
\label{tab:seven_dataset_audit}
\resizebox{\textwidth}{!}{%
\begin{tabular}{l r c c c c r r l}
\toprule
Dataset & $N$ & $I_{\mathrm{lin}}$ & $I_{\mathrm{gbt}}$ & Beat & Perm\% & iid $R^2$ & Group $R^2$ & Profile \\
\midrule
OULAD & 32,593 & 0.010 & 0.008 & 1.00 & 1.00 & 0.471 & 0.410 & Strong (4/4) \\
Dropout & 3,630 & 0.021 & 0.019 & 1.00 & 1.00 & 0.651 & 0.527 & Strong (4/4) \\
xAPI-Edu & 480 & 0.124 & 0.102 & 1.00 & 1.00 & 0.614 & 0.484 & Strong (4/4) \\
Entrance Exam & 666 & 0.117 & 0.115 & 1.00 & 1.00 & 0.438 & $-0.060^{\dagger}$ & Mostly Passing (3/4) \\
UCI Student & 649 & 0.363 & 0.406 & 1.00 & 1.00 & 0.242 & $-0.097^{\dagger}$ & Mostly Passing (3/4) \\
Higher Ed & 145 & 0.895 & 0.226 & 0.83 & 0.73 & 0.041 & $-8.79^{\dagger}$ & Partial (1/4) \\
MM-TBA & 186 & 2.212 & 4.604 & 0.80 & 0.23 & $-0.067$\textsuperscript{\ddag} & N/A & Fragile (0/4) \\
\bottomrule
\end{tabular}}%
\par\vspace{4pt}
\begin{flushleft}\footnotesize
$^{\dagger}$ Linear-model group $R^2$ with group-identifier features excluded from the feature matrix before leave-one-group-out training (see Section~\ref{sec:models_eval}). Regularized and ensemble models often perform better (see Table~\ref{tab:group_holdout_detail}).\par
$^{\ddag}$ MM-TBA shows a positive MAE gap over the mean baseline ($\Delta_{\mathrm{MAE}} = 0.040$, beat rate $0.80$) but a negative iid $R^2$, i.e., its average squared-error fit is worse than the variance-only mean baseline. The two metrics disagree because $\Delta_{\mathrm{MAE}}$ rewards small absolute reductions in median-error scale while $R^2$ penalises large residuals; both signals support the Fragile profile.\par
The final metadata-adequacy Pass/Fail judgment uses the linear-model group $R^2$ (retention $\geq 50\%$ of linear iid $R^2$) so that the audit flags the weakest link in the benchmark; the detailed breakdown in Table~\ref{tab:group_holdout_detail} further distinguishes structurally invariant collapse (where all models fail, as with UCI Student and Higher Ed) from model-dependent failure that is partly remediable by regularization or ensemble methods (as with Entrance Exam).\par
Perm\% 95\% Clopper--Pearson CIs: 0.23 $\to$ [0.10, 0.42]; 0.73 $\to$ [0.54, 0.88]; 1.00 $\to$ [0.88, 1.00] ($n=30$ splits).\par
Sources: MM-TBA \citep{Huang2025}; UCI Student \citep{Cortez2008}; OULAD \citep{Kuzilek2017}; xAPI-Edu \citep{Amrieh2016}; Entrance Exam \citep{Bora2021}; Dropout \citep{Realinho2022}; Higher Ed \citep{Yilmaz2020}.
\end{flushleft}
\end{table*}

Three of six testable educational datasets show evidence of cross-group generalization failure after correcting for group-identifier features in the feature matrix (see Section~\ref{sec:models_eval}). Linear-model group $R^2$ is substantially below iid $R^2$ for Entrance Exam (retention $-14\%$), UCI Student (retention $-40\%$), and Higher Ed (catastrophic $R^2 = -8.79$), while MM-TBA cannot be tested due to missing grouping metadata. The nature of the failure differs: UCI Student and Higher Ed exhibit structural fragility (all model families collapse), whereas Entrance Exam's failure is linear-model-specific and partly remediable---its RF group $R^2$ reaches approximately $0.32$. xAPI-Edu passes (retention 79\%), confirming that its original failure was a feature-encoding artifact. Three datasets---OULAD, Dropout, and xAPI-Edu---survive group-aware holdout with positive $R^2$ for the linear model after this correction.

\begin{figure}[!t]
\centering
\includegraphics[width=0.95\textwidth]{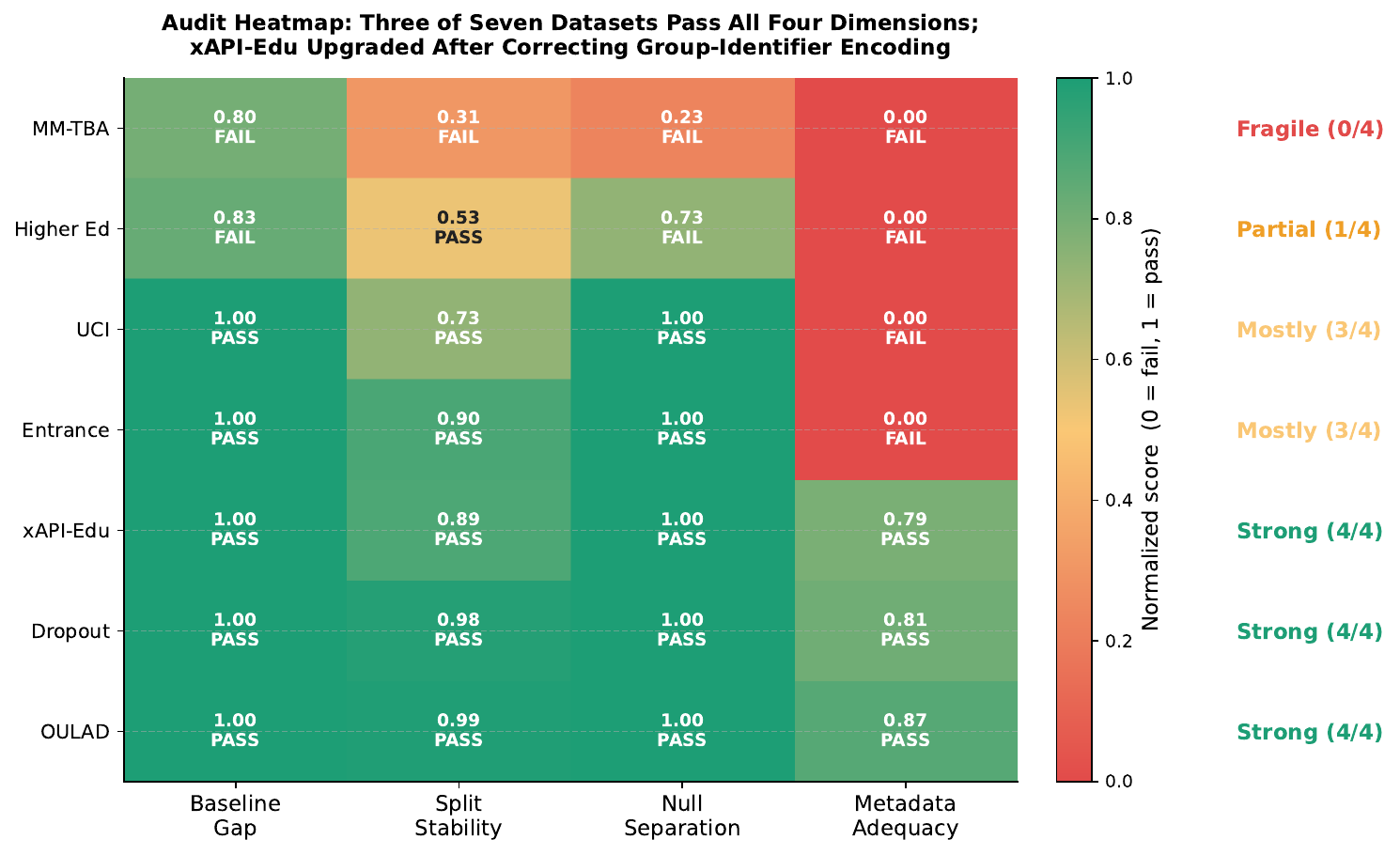}
\caption{Audit summary heatmap: four-dimension normalised scores for the seven datasets. Cells show the score (0--1) and the corresponding PASS/FAIL judgement under the operational thresholds (beat rate $\geq 0.90$; $1 - I/(1+I) \geq 0.50$, i.e.\ $I < 1$; permutation significance $\geq 0.80$; group-holdout retention $\geq 0.50$ using the \emph{linear} model). The right margin lists the assigned readiness profile. Four of seven datasets fail metadata adequacy (including MM-TBA, which lacks grouping metadata); OULAD, Dropout, and xAPI-Edu pass all four dimensions.}
\label{fig:audit_summary}
\end{figure}

A negative group-holdout $R^2$ has a concrete interpretation: the model's predictions for students in the held-out group are \textit{worse than simply predicting the overall mean}---i.e., the model has learned group-specific patterns that actively mislead when applied to unseen groups. For educational deployment, this means a model trained on data from some schools (or courses, or cohorts) could systematically mispredict outcomes for students in new institutional contexts. Fig~\ref{fig:audit_summary} summarises the four-dimension audit scores side by side, confirming the graded nature of the failure pattern.

\begin{figure}[!t]
\centering
\includegraphics[width=0.95\textwidth]{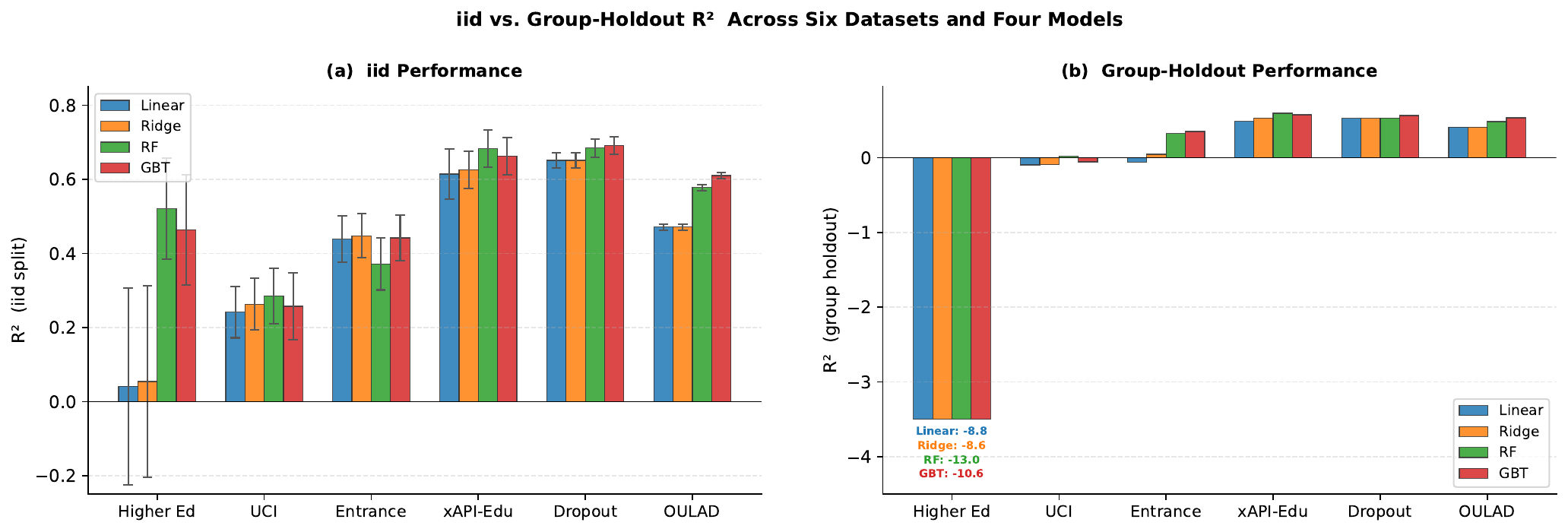}
\caption{Comparison of iid $R^2$ (left panel) and group-holdout $R^2$ (right panel) across six datasets and four models. Datasets are ordered by audit profile. The contrast between iid and group-holdout performance is largest for Higher Ed, where all models collapse under group holdout. MM-TBA is omitted because it has no grouping metadata.}
\label{fig:iid_vs_group}
\end{figure}

\Needspace{4\baselineskip}
Group-holdout results reveal an important model-dependent pattern (see Fig~\ref{fig:iid_vs_group} for a visual comparison of iid vs.\ group-holdout $R^2$). While unregularized linear regression collapses under group holdout on several datasets due to overfitting (especially on high-dimensional one-hot-encoded features), regularized and ensemble models often retain positive $R^2$.  Table~\ref{tab:group_holdout_detail} shows this pattern.

\begin{table*}[htbp]
\centering
\caption{Group-holdout $R^2$ by model across datasets where group-aware holdout is available. Regularized models (ridge) and ensemble methods often survive group holdout where unregularized linear regression collapses, indicating that the group-level confounding is partly a function of model capacity matching on high-dimensional features.}
\label{tab:group_holdout_detail}
\begin{tabular}{l r r r r}
\toprule
Dataset & Linear & Ridge & RF & GBT \\
\midrule
OULAD & 0.410 & 0.410 & 0.482 & 0.533 \\
Dropout & 0.527 & 0.527 & 0.531 & 0.564 \\
xAPI-Edu & 0.484 & 0.532 & 0.594 & 0.573 \\
Entrance Exam & $-0.060$ & 0.047 & 0.322 & 0.351 \\
UCI Student & $-0.097$ & $-0.094$ & 0.018 & $-0.057$ \\
Higher Ed & $-8.79$ & $-8.58$ & $-13.0$ & $-10.6$ \\
\bottomrule
\end{tabular}
\end{table*}

This pattern reveals a graded structure that becomes clearer after removing group-identifier features from the group-holdout feature matrix (see Section~\ref{sec:models_eval}). On xAPI-Edu, the linear model now achieves positive group $R^2$ (0.484), comparable to ridge (0.532) and RF (0.594), confirming that its previous apparent collapse was an artifact of Topic being encoded as a one-hot feature. On Entrance Exam, the linear model remains near zero ($-0.060$) while ensemble models retain moderate performance ($R^2 \approx 0.32$--$0.35$). On UCI Student, even after removing school features, all models yield near-zero group $R^2$, confirming genuine school-level confounding beyond feature leakage. On Higher Ed, all models collapse catastrophically, confirming that the small dataset ($N=145$) with 9 course groups cannot support cross-group generalization.

Group-holdout $R^2$ collapse does not uniquely diagnose confounding; it may also reflect legitimate distributional shift when held-out groups differ systematically from the training groups. To distinguish these cases, the following diagnostic sequence is recommended (see also Discussion). When all model types collapse (as with UCI Student and Higher Ed), comparing group-level feature distributions can reveal whether large distributional gaps point toward cross-context heterogeneity rather than leakage. When only unregularized linear models collapse while ridge or ensemble models recover (as with Entrance Exam), overfitting on high-dimensional features is the more parsimonious explanation.

The supplementary analysis quantifies the across-group variability of these estimates (Supplementary Table~S4). For UCI Student, the linear-model group-holdout $R^2$ carries a standard deviation of 0.109 across its 2 eligible schools (range: $-0.206$ to 0.012), confirming uniform rather than group-specific collapse. For OULAD, the across-cohort SD is 0.162 across 22 cohorts (range: 0.041 to 0.654), consistent with positive---though variable---cross-group generalization. For Higher Ed (4 eligible courses), the large SD of 6.39 (range: $-18.65$ to $-0.975$) reflects the high heterogeneity of collapse severity across courses.

A dedicated train-test $R^2$ gap analysis (Supplementary Table~S3) further separates overfitting from structural fragility. On UCI Student, the linear model's mean train-test gap is 0.105 (train $R^2 = 0.347$, test $R^2 = 0.242$), compared to an iid-to-group collapse from 0.242 to $-0.097$. This contrast confirms that the group-holdout failure is not attributable to ordinary overfitting---the model generalizes adequately to iid held-out samples from the same distribution but fails specifically on unseen school groups. On OULAD, both the train-test gap (0.008) and the iid-to-group gap (0.061) remain small. On Higher Ed, the train-test gap is larger (0.552) and the group-holdout collapse is catastrophic ($-8.79$), indicating that overfitting and cross-group fragility are jointly present.

\subsection{Per-Dataset Results}

The three Strong datasets confirm that the audit cleanly passes structurally sound data. \textbf{OULAD} achieves $R^2 = 0.471$, $I = 0.010$, permutation significance $= 1.00$, and group-holdout $R^2 = 0.410$ across 22 cohorts. \textbf{Dropout} achieves $R^2 = 0.651$, $I = 0.021$, and group-holdout $R^2 = 0.527$ across 16 of 17 course programs (one program excluded for $n < 10$). \textbf{xAPI-Edu} achieves $R^2 = 0.614$, $I = 0.124$, and group-holdout $R^2 = 0.484$ (retention 79\%) across 12 topics after correcting for the group-column encoding artifact described in Section~\ref{sec:models_eval}.

\textbf{xAPI-Edu} (Strong) passes all four dimensions. Its linear-model group-holdout $R^2$ of 0.484 (retention 79\%) confirms that the apparent collapse in prior analyses was an artifact of Topic being included as a one-hot feature; with Topic features excluded, all models show positive group $R^2$ (0.48--0.59). \textbf{Entrance Exam} (Mostly Passing) passes the first three dimensions but fails linear-model group holdout ($R^2 = -0.060$, retention $-14\%$), while ensemble models recover moderate performance (RF $R^2 \approx 0.32$).

\textbf{UCI Student} (Mostly Passing) shows persistent cross-school confounding: iid $R^2 = 0.242$ drops to $-0.097$ under school-level holdout (retention $-40\%$), with RF ($0.018$) and GBT ($-0.057$) also near zero even after school features are excluded from the group-holdout feature matrix. \textbf{Higher Ed} (Partial) is borderline on baseline gap (beat rate $= 0.83$) and instability ($I = 0.90$), lacks consistent null separation (Perm\% $= 0.73$), and shows catastrophic group-holdout collapse for all models (linear $-8.79$, RF $-13.0$).

\begin{figure}[t]
\centering
\includegraphics[width=0.95\textwidth]{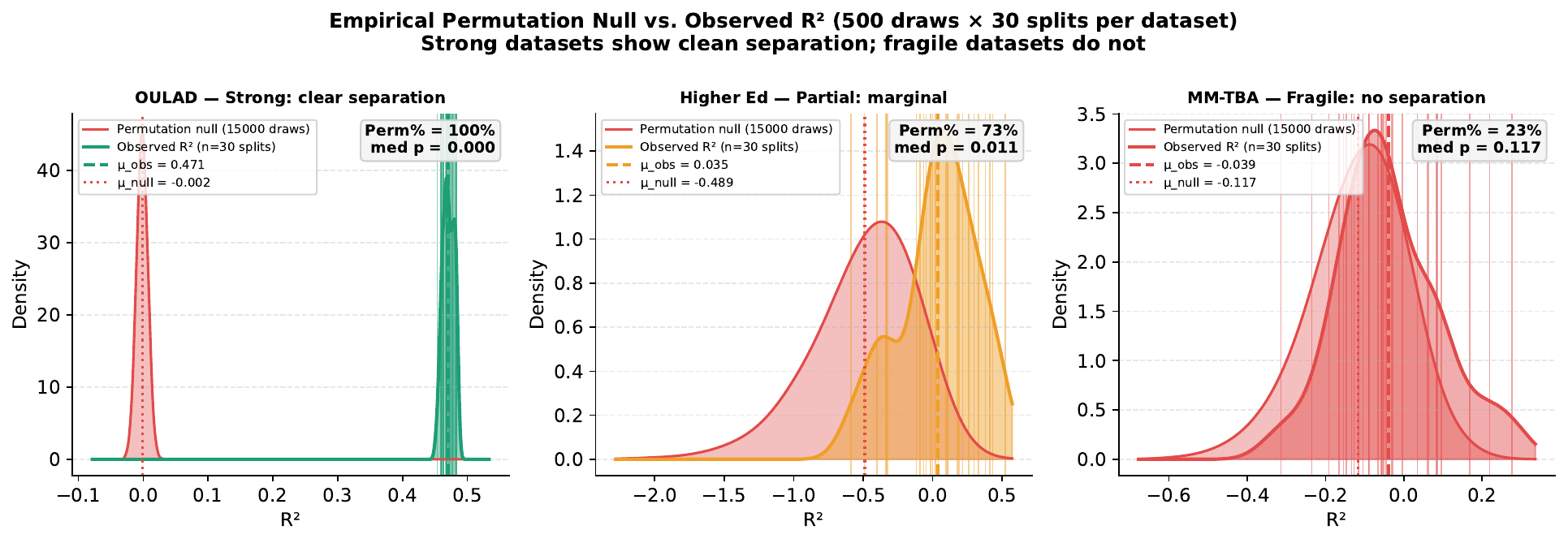}
\caption{Empirical permutation null vs.\ observed linear $R^2$ for three representative datasets. The red density is the pooled empirical null distribution from 500 permutation draws on each of 30 random 80/20 splits (15{,}000 values total per dataset); the coloured density is the observed-label $R^2$ across the same 30 splits, with vertical ticks marking individual splits and dashed/dotted lines showing the means. OULAD shows clean separation (Perm\% $= 100\%$), Higher Ed is borderline but below the pass threshold (Perm\% $= 73\%$), and MM-TBA shows substantial overlap (Perm\% $= 23\%$, median $p = 0.117$).}
\label{fig:null}
\end{figure}

\textbf{MM-TBA} (Fragile) fails all four dimensions: $I = 2.21$, beat rate $= 0.80$, permutation significance $= 0.23$, and no grouping metadata. The linear model's iid $R^2 = -0.067$ is itself negative, meaning the model's squared-error fit is worse than the variance-only mean baseline even though the MAE-based beat rate is $0.80$. This metric disagreement is diagnostic of fragility rather than a contradiction. GBT worsens the picture ($I = 4.60$, beat rate $= 0.45$). Fig~\ref{fig:null} contrasts the empirical permutation null (500 draws $\times$ 30 splits per dataset) with the observed linear $R^2$ for OULAD, Higher Ed, and MM-TBA, illustrating how the null-separation dimension distinguishes Strong, borderline, and Fragile cases.

\subsection{Cross-Dataset Structural Patterns and iid-to-Group Collapse}

\begin{figure}[!t]
\centering
\includegraphics[width=0.95\textwidth]{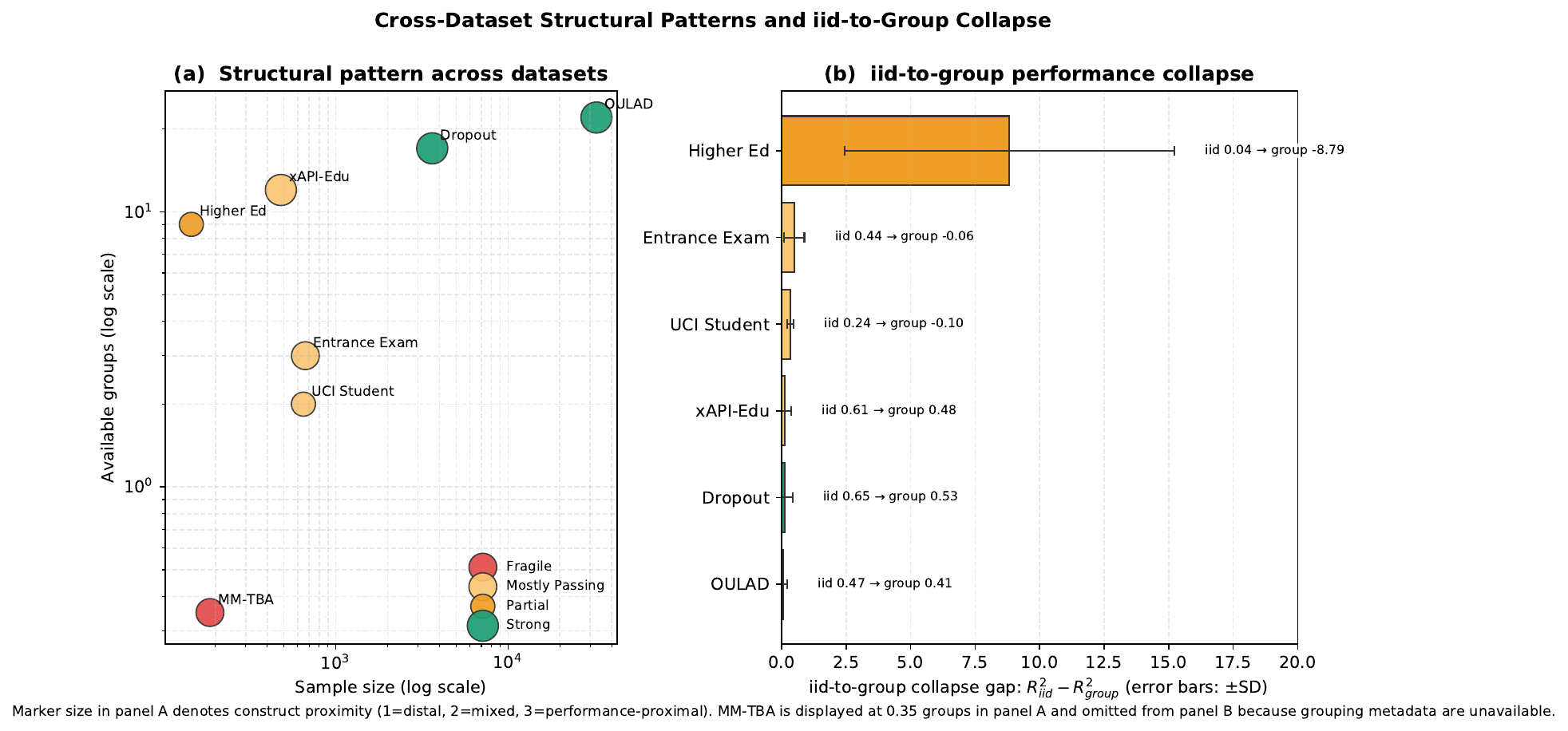}
\caption{Exploratory cross-dataset structural pattern and the practical cost of skipping group-aware audit. Panel A places the seven datasets by sample size and available grouping structure; marker size denotes construct proximity (1 = distal background/context variables, 2 = mixed features, 3 = performance-proximal behavioral or administrative traces), and colour denotes readiness profile. MM-TBA is displayed at 0.35 groups because no grouping metadata are available. Panel B reports the iid-to-group collapse gap, $R^2_{iid} - R^2_{group}$, for datasets with grouping metadata. Strong datasets show small collapse gaps (all $\leq 0.13$), whereas Higher Ed (group $R^2 = -8.79$), Entrance Exam, and UCI Student show larger gaps.}
\label{fig:structural_patterns}
\end{figure}

Because the audit includes only seven datasets, the cross-dataset comparison should be interpreted as an exploratory structural pattern analysis rather than a formal meta-regression. Even with this caveat, the profiles show an interpretable descriptive pattern rather than an arbitrary ordering (Fig~\ref{fig:structural_patterns}). The three Strong datasets contain many usable groups for holdout validation and use features close to the outcome construct: OULAD combines assessment and VLE behavioral traces with a pass/fail outcome; Dropout contains administrative and performance-proximal records; xAPI-Edu includes behavioral traces from an online learning platform. By contrast, the weakest profiles occur in small releases, sparsely grouped releases, or releases with distal or weakly documented feature-target links. Higher Ed has only $N=145$ observations and fails three of four dimensions; MM-TBA lacks grouping metadata and has a target derived from an automated rubric pipeline; UCI Student has only two school groups, making leave-one-school-out validation a severe cross-context transfer test.

The same comparison quantifies the practical cost of relying only on random-split performance. For the Strong datasets, the iid-to-group collapse gap is small (xAPI-Edu: 0.614 to 0.484, gap 0.130; OULAD: 0.062; Dropout: 0.125), meaning that random-split performance gives a reasonably faithful picture of group-aware performance. For the datasets with metadata-adequacy failures, the gap is larger: Entrance Exam drops from iid $R^2 = 0.438$ to group $R^2 = -0.060$ (linear model; ensemble models partly recover), Higher Ed from $0.041$ to $-8.79$ (all models collapse), and UCI Student from $0.242$ to $-0.097$ (all models near zero). In practical terms, a model that appears useful under random splitting can become worse than a mean-only predictor when transferred to unseen schools, courses, topics, or education boards. This collapse is the operational reason the audit is framed as a pre-modeling quality gate rather than an optional reporting add-on.

\subsection{Model-Complexity Ablation}

\begin{figure}[!t]
\centering
\includegraphics[width=0.95\textwidth]{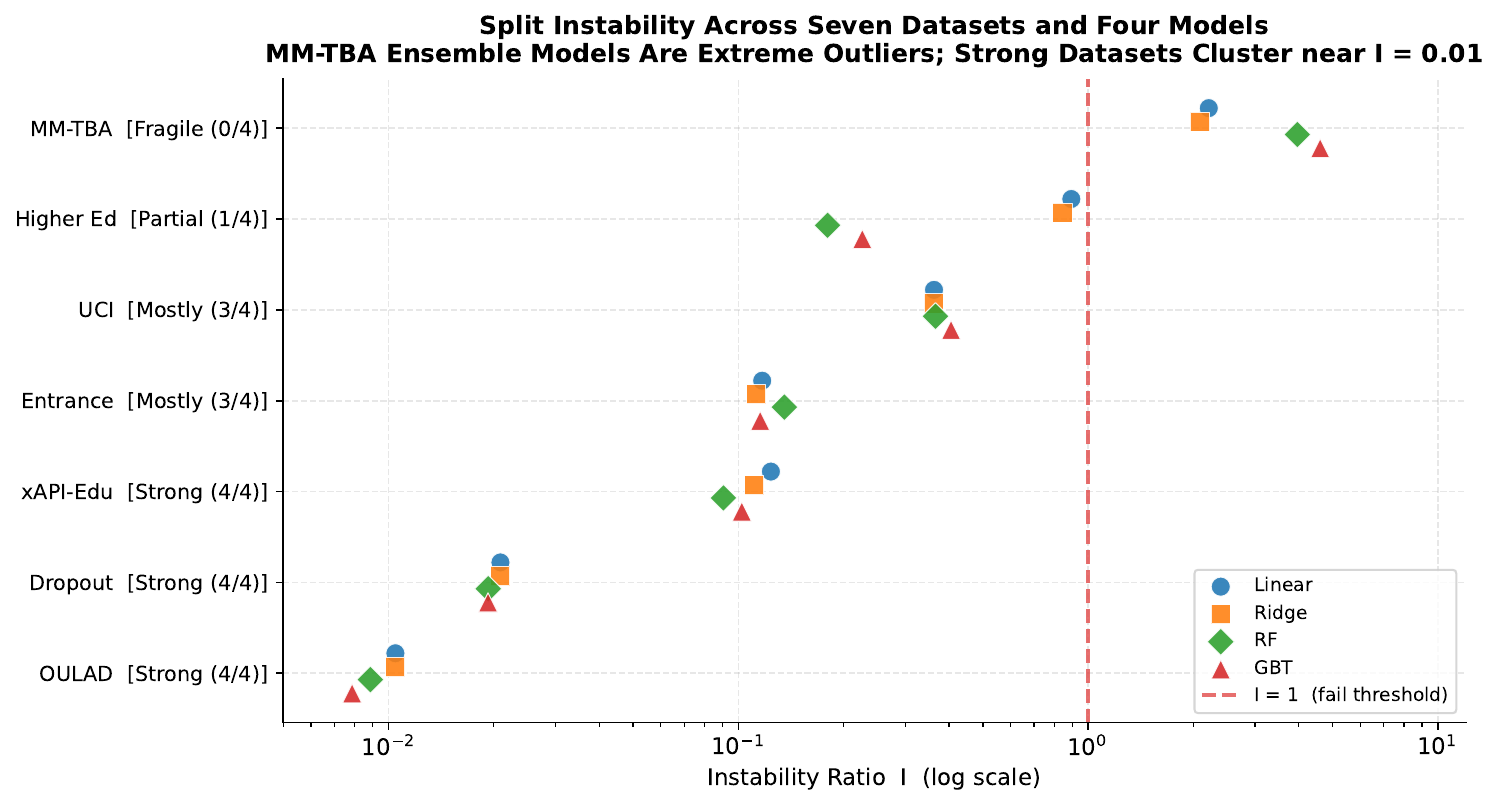}
\caption{Instability ratio ($I$) for all seven datasets and four models (log scale). The dashed red line marks $I = 1$, the failure threshold where between-split variance equals the mean advantage over baseline. MM-TBA's Gradient Boosting ($I = 4.6$) and Random Forest ($I = 4.0$) are extreme outliers; structurally sound datasets (OULAD, Dropout) cluster near $I = 0.01$.}
\label{fig:instability_strip}
\end{figure}

The full model-complexity ablation (detailed results in Supplementary Table S1) reports instability ratio and beat rate for linear, ridge, RF, and GBT models across all seven datasets. The key patterns are as follows. On fragile data (MM-TBA), ensemble models amplify instability ($I$ rising from 2.2 to 4.6). On partially passing data (Higher Ed), ensemble models reduce iid instability ($I$ dropping from 0.90 to 0.18--0.23) and improve iid $R^2$ from 0.04 to 0.52, but collapse equally under group holdout. On structurally sound data (OULAD, Dropout), ensemble models yield genuine gains (e.g., OULAD RF MAE $0.193$ vs. linear $0.296$) without introducing instability. Fig~\ref{fig:instability_strip} plots the instability ratio across all datasets and models on a logarithmic scale, making the $I = 1$ failure boundary visually salient.

\subsection{Classification-Metric Sensitivity Check}

To verify that the regression-framing audit conclusions are not an artifact of metric choice, the four datasets with binary or ordinal targets were re-evaluated under classification-appropriate metrics (Table~\ref{tab:classification_sensitivity}).

Classification-metric results confirm the regression-framing conclusions (Table~\ref{tab:classification_sensitivity}). OULAD, Dropout, and xAPI-Edu retain Strong profiles (iid accuracy 0.74--0.91, $I_{\mathrm{acc}} \leq 0.14$, group-holdout AUC $\geq 0.86$). Entrance Exam shows the weakest performance (Logistic iid accuracy 51.2\% on a 4-class task with majority baseline 30.5\%, group-holdout AUC $= 0.68$--$0.73$), consistent with the regression-framing assessment of moderate but fragile signal; the low absolute accuracy partly reflects the inherent difficulty of 4-class ordinal prediction.

\begin{table*}[!ht]
\centering
\caption{Classification-metric sensitivity check on the four datasets with binary or ordinal targets. Accuracy and AUC-ROC are reported for each classifier under iid 100-split resampling and group-aware holdout. $I_{\mathrm{acc}}$ is the accuracy-based instability ratio. Group-holdout results for xAPI-Edu and Entrance Exam use the corrected feature matrix with group-identifier columns excluded (see Section~\ref{sec:models_eval}).}
\label{tab:classification_sensitivity}
\resizebox{\textwidth}{!}{%
\begin{tabular}{l l c c c c c}
\toprule
Dataset & Classifier & iid Acc (mean $\pm$ SD) & iid AUC & $I_{\mathrm{acc}}$ & Group Acc (mean) & Group AUC \\
\midrule
OULAD & Logistic & $0.837 \pm 0.004$ & $0.910$ & $0.013$ & $0.817$ & $0.920$ \\
(binary) & RF & $0.849 \pm 0.004$ & $0.922$ & $0.012$ & $0.825$ & $0.926$ \\
 & GBT & $0.861 \pm 0.004$ & $0.938$ & $0.012$ & $0.842$ & $0.939$ \\
\midrule
Dropout & Logistic & $0.912 \pm 0.009$ & $0.953$ & $0.031$ & $0.897$ & $0.943$ \\
(binary) & RF & $0.907 \pm 0.010$ & $0.952$ & $0.035$ & $0.885$ & $0.928$ \\
 & GBT & $0.909 \pm 0.010$ & $0.954$ & $0.033$ & $0.886$ & $0.938$ \\
\midrule
xAPI-Edu & Logistic & $0.736 \pm 0.037$ & $0.869$ & $0.131$ & $0.706$ & $0.856$ \\
(3-class) & RF & $0.795 \pm 0.035$ & $0.916$ & $0.102$ & $0.730$ & $0.896$ \\
 & GBT & $0.757 \pm 0.043$ & $0.883$ & $0.139$ & $0.672$ & $0.857$ \\
\midrule
Entrance Exam & Logistic & $0.512 \pm 0.039$ & $0.762$ & $0.188$ & $0.390$ & $0.698$ \\
(4-class) & RF & $0.487 \pm 0.038$ & $0.725$ & $0.207$ & $0.416$ & $0.682$ \\
 & GBT & $0.502 \pm 0.040$ & $0.749$ & $0.203$ & $0.493$ & $0.734$ \\
\bottomrule
\end{tabular}}%
\end{table*}

\FloatBarrier
\subsection{Feature-Attribution Stability}
\label{sec:feat_attribution}

To characterise the cross-split stability of feature attributions, linear-model standardized coefficients and Random Forest feature importances (impurity-based, using scikit-learn's default \texttt{feature\_importances\_}) are tracked across 100 repeated 80/20 splits for UCI Student, Higher Ed, and OULAD. These three datasets span three of the four readiness profiles (Strong, Mostly Passing, and Partial). The Fragile profile is not represented in this exploratory analysis. (Supplementary Tables~S5--S6).

On OULAD, the top-10 linear coefficients show very low cross-split standard deviations (SD $\leq 0.007$) with zero sign flips, indicating consistent directional relationships. Random Forest importances are similarly stable, with coefficients of variation (CV) below 0.12 for the top-10 features. On UCI Student, coefficient SDs are larger (up to 0.097 for f9, which shows a 3\% sign-flip rate), and RF importance CVs range from 0.08 to 0.20, reflecting greater uncertainty about feature rankings. On Higher Ed, the pattern is most pronounced: coefficient SDs reach 0.134 (f0), and RF importance CVs range from 0.18 to 0.41, meaning that feature rankings for the smallest dataset ($N=145$) vary substantially across splits.

Split-level attribution instability is an exploratory diagnostic: when a model's feature weights vary substantially across random resamples of the same dataset, the resulting attribution estimates may be sensitive to the sampled partition. Because this analysis is conducted only under IID splits and does not directly assess attribution stability under group-aware holdout, the observed patterns provide exploratory associations that are consistent with---but do not validate---the audit profiles. Within the three datasets analysed, the descriptive patterns align with the audit classifications: on OULAD (Strong), feature-attribution is stable; on UCI Student (Mostly Passing) and Higher Ed (Partial), feature-attribution rankings are more variable across IID splits. A cross-model comparison (Supplementary Table~S8) is also consistent with this pattern: on OULAD, the Spearman rank correlation between linear absolute coefficients and RF importances is high and stable ($\rho_{\mathrm{top10}} = 0.817$, SD 0.065 across 100 splits); on UCI Student and Higher Ed, the agreement is lower ($0.516 \pm 0.242$ and $0.498 \pm 0.245$, respectively) and substantially more variable, suggesting that attribution rankings from the Partial and Mostly Passing datasets are more model-dependent as well as split-dependent. Even on OULAD, the all-feature agreement is strong ($\rho_{\mathrm{all}} = 0.752$), indicating that linear and nonlinear models capture partially different feature relationships beyond the top-ranked predictors. Together, these results are descriptively consistent with the audit profiles across the three analysed datasets: OULAD (Strong) maintains stable attribution patterns across splits, while UCI Student (Mostly Passing) and Higher Ed (Partial) exhibit more split-dependent rankings. These exploratory associations do not, by themselves, establish a causal link between IID attribution instability and cross-group fragility.

\FloatBarrier
\section{Discussion}

As an exploratory analysis of seven datasets, these findings should not be over-generalized to the full population of educational prediction benchmarks. Within this scope, the results answer the study's three research questions. First, benchmark reliability is uncommon: three of the seven public datasets pass all four checks. Second, the dominant failure mode is cross-group fragility rather than weak iid performance alone. Third, these conclusions remain substantively unchanged when model complexity and classification-appropriate metrics are varied.

\subsection{Findings}

Four findings follow from the seven-dataset audit.

\textbf{Finding 1: Cross-group generalization failure is pervasive but graded.} After correcting for group-identifier features in the holdout matrix, two datasets exhibit structural fragility: Higher Ed shows model-invariant catastrophic collapse ($R^2 < -8$ for all models) and UCI Student shows model-invariant near-zero holdout performance (linear $-0.097$, RF $0.018$, GBT $-0.057$). Entrance Exam presents a distinct pattern---linear-model-only failure ($R^2 = -0.060$, retention $-14\%$) that is partly remediable by ensemble models (RF $R^2 \approx 0.32$), indicating that its metadata-adequacy failure is model-dependent rather than structurally invariant. MM-TBA lacks grouping metadata entirely, failing the dimension by default. One case---xAPI-Edu---passes after removing group-identifier features from the holdout matrix, confirming that its original failure was an artifact of encoding Topic as a one-hot predictor. This distinction matters for practitioners: xAPI-Edu is usable with proper leakage prevention, UCI Student requires stronger cross-school evidence, and Higher Ed requires more courses and larger samples per course, while Entrance Exam's group-level signal is recoverable through regularization or ensemble methods.

\textbf{Finding 2: Structural fragility persists across model families, while model-dependent failures are partly remediable.} On MM-TBA (structurally fragile), upgrading from linear regression ($I = 2.2$) to Gradient Boosting ($I = 4.6$) doubles the instability. On Higher Ed (structurally fragile), RF improves iid $R^2$ from 0.04 to 0.52 but collapses equally under group holdout ($R^2 = -13.0$). On Entrance Exam (model-dependent failure), ensemble models substantially recover group-holdout performance (RF $R^2 \approx 0.32$ vs.\ linear $-0.060$), confirming that its metadata-adequacy failure is partly remediable through modeling choices. On structurally sound datasets, ensemble models yield genuine gains without introducing instability (see Fig~\ref{fig:instability_strip}).

\textbf{Finding 3: Reliability profiles follow interpretable structural patterns.} The seven datasets span a structured continuum from Fragile (0/4) through Partial (1/4) and Mostly Passing (3/4) to Strong (4/4). This gradient should be interpreted cautiously because seven datasets are insufficient for formal meta-regression. The descriptive pattern is nonetheless informative: stronger profiles cluster in larger datasets with richer grouping structure and more performance-proximal features, whereas weaker profiles cluster in smaller, sparsely grouped, or weakly documented releases. OULAD's strong profile is partly expected because its behavioral-performance features (assessment scores, VLE clicks) are definitionally close to the binary pass/fail outcome. An instructive contrast is UCI Student vs. xAPI-Edu: despite similar sample sizes ($N = 649$ vs. $480$) and respective iid $R^2$ ($0.24$ vs. $0.61$), they diverge under group holdout---UCI shows persistent cross-school confounding ($R^2 \approx 0.00$ even after excluding school features), while xAPI-Edu passes with all models ($R^2 = 0.48$--$0.59$). The divergence traces to the nature of the grouping structure: UCI's 2 schools (GP: $n=423$; MS: $n=226$) represent distinct institutional contexts with large distributional gaps, whereas xAPI-Edu's 12 topics share a common platform and student population. With only 2 groups, UCI's leave-one-group-out holdout effectively trains on one school and tests on another, making the evaluation maximally sensitive to between-group heterogeneity. This illustrates that group-holdout severity depends on both the number and heterogeneity of groups, not merely on sample size.

\textbf{Finding 4: Random-split performance can overstate deployable signal.} The iid-to-group collapse gap is small for Strong datasets but larger for datasets with metadata-adequacy failures. UCI Student is a clear case: iid evaluation suggests modest signal ($R^2 = 0.242$), yet school-level holdout drops to $R^2 = -0.097$. Higher Ed (from $0.041$ to $-8.79$) shows catastrophic collapse across all models. Entrance Exam shows a narrower gap ($0.438$ to $-0.060$) that is partly closed by ensemble models. These gaps convert an abstract reliability problem into a practical one: without group-aware auditing, a benchmark can encourage model selection decisions that would fail when transferred to a new educational context.

The practical value of the profiles is that each one implies a different remediation path. \textbf{Fragile} datasets should not be used for benchmark claims until the target construction, grouping metadata, and feature provenance are revisited; in MM-TBA, this would require teacher- or classroom-level identifiers and independent evidence that rubric targets are not artifacts of the scoring pipeline. \textbf{Partial} datasets require additional data or broader group coverage before cross-context claims are made; Higher Ed would benefit from more courses and larger held-out course samples. \textbf{Mostly Passing} datasets can support limited benchmark use if the reported model is paired with regularization-aware validation and explicit group-holdout diagnostics; Entrance Exam is an example where model choice partly repairs linear collapse, whereas UCI Student requires stronger cross-school evidence. \textbf{Strong} datasets support cautious benchmark claims, but even OULAD, Dropout, and xAPI-Edu should still report group-aware performance when used for deployment-oriented comparisons.

These findings extend prior work on benchmark reliability \citep{Northcutt2021,Recht2019,Bouthillier2021} by showing that group-level confounding and missing provenance metadata impose constraints in educational prediction comparable to—and often exceeding—those of label quality or split variation in vision benchmarks. The results are also consistent with the data documentation literature \citep{Gebru2021,Sambasivan2021}, which identifies upstream quality failures as downstream benchmark threats. The contribution lies in integrating these concerns into a unified four-dimension pre-modeling audit protocol and demonstrating, across seven public releases, that their combined effect renders most of these benchmarks structurally unreliable for strong benchmark claims.

The audit findings carry direct socio-technical implications for educational deployment. The UCI school-level fragility means a model deployed across schools could systematically mispredict outcomes. The MM-TBA instability means the same teacher could be rated differently depending on which random split was used. The iid-to-group collapse analysis further shows that the risk is not merely statistical: random-split performance can substantially overstate how a model will behave in unseen educational groups. The audit provides a minimum-viable quality gate before educational prediction models influence decisions about teachers or students.

These dimensions also connect the audit protocol to established educational measurement theory. The instability ratio $I$ operationalizes a concern central to Generalizability Theory \citep{Brennan2001}: when $I \geq 1$, the benchmark conclusion is not generalizable across train-test partitioning. The group-holdout dimension operationalizes a Messick~\citep{Messick1995}-style validity requirement: a dataset whose signal collapses under group-aware evaluation lacks generalizability evidence. An important caveat is that group-holdout collapse does not uniquely diagnose confounding; it may also reflect legitimate distributional shift. Distinguishing the two requires domain knowledge that the audit flags the need for but does not itself provide. As a practical heuristic, the following diagnostic sequence is suggested. When all model types collapse under group holdout (as with UCI Student and Higher Ed), the first step is to compare group-level feature distributions---large distributional gaps between groups point toward genuine cross-context heterogeneity rather than statistical confounding. When only unregularized linear models collapse while ridge or ensemble models recover (as with Entrance Exam), overfitting on high-dimensional features is the more parsimonious explanation, and regularization may suffice. When group counts are very small ($k \leq 3$), leave-one-group-out holdout inherently functions as cross-distribution extrapolation, and results should be interpreted as stress tests of out-of-distribution robustness rather than direct evidence of latent confounders.

A persistent practice gap motivates broader adoption of these diagnostics. Educational prediction papers rarely report split-instability diagnostics, permutation nulls, or group-aware holdout results. Applying these diagnostics systematically reveals evaluation fragility that headline metrics conceal. Related frameworks---Model Cards \citep{Mitchell2019} and Dataset Nutrition Labels \citep{Holland2020}---address complementary concerns; the protocol developed here adds a computational stress-test layer. For instance, \citet{Sasmaz2026} applied a multi-model framework with stratified $k$-fold cross-validation to PISA 2022 data but did not report group-aware holdout results, illustrating the gap between current practice and the audit standards proposed here. That OULAD supports post-modeling interpretability analyses as demanding as counterfactual explanation benchmarking \citep{Cavus2024ce,Cavus2024benchmark} is consistent with its Strong (4/4) audit profile: datasets that fail group-aware holdout are unlikely to provide the stable distributional foundation needed for reliable post-modeling interpretability work.

\subsection{Limitations}

The seven-dataset gradient should not be over-interpreted as evidence that the audit procedure has inherent discriminative calibration, because it is partly confounded with sample size, feature-target proximity, and task-construct differences. The operational thresholds ($I \geq 1$, group-holdout $R^2$ retention $\geq 50\%$) are empirically motivated but not yet calibrated across a large benchmark corpus. For example, Higher Ed's linear-model beat rate of 0.83 places it in the borderline zone between pass (0.90) and fail (0.80): at a pass threshold of 0.80 it would pass baseline gap, but at 0.90 it fails. Similarly, a 40\% (rather than 50\%) group-holdout retention threshold would not change any current assignment but could affect future datasets near the boundary. Until the thresholds are calibrated on a larger benchmark corpus, practitioners should treat them as informed defaults rather than universal cutoffs.

The permutation-null dimension uses 30 of the 100 random splits per dataset rather than all 100, so Perm\% estimates still carry binomial uncertainty, especially for borderline datasets. This uncertainty does not affect datasets with complete null separation, but it does matter for cases near the 80\% decision threshold. Future large-scale benchmark audits could run permutation tests on all repeated splits or use adaptive stopping rules that spend more computation on borderline datasets.

The group-holdout results are sensitive to the number and size of groups: UCI has only 2 schools, Entrance Exam has 3 boards, while Dropout has 17 programs---making cross-dataset comparisons of group-holdout severity approximate. The four audit dimensions are currently given equal weight in assigning profile labels (Fragile through Strong); in practice, domain context may warrant differential emphasis---for example, metadata adequacy may be more consequential than split instability for high-stakes deployment decisions.

The seven audited datasets span a range of geographical and institutional contexts---European (OULAD, Higher Ed, UCI Student, Dropout), Jordanian (xAPI-Edu), Indian (Entrance Exam), and Chinese (MM-TBA) educational systems---which provides some cross-regional breadth; nevertheless, contexts such as sub-Saharan Africa, North America, and Latin America are not represented, and generalization to those settings remains untested. The audit uses linear baselines and two ensemble methods; other model families (e.g., deep learning) might yield different instability patterns on some datasets, though the algorithm-invariance finding across four model types suggests this is unlikely to change the main conclusions. A stronger next-stage study should include datasets with richer grouping metadata (e.g., teacher-level or classroom-level identifiers) and, where available, paired evidence on agreement between automated and human scoring.

The four audit dimensions target pre-modeling statistical reliability rather than post-modeling properties such as fairness, algorithmic robustness, or explainability. These omitted dimensions are complementary rather than competing: Model Cards \citep{Mitchell2019} and Datasheets for Datasets \citep{Gebru2021} provide frameworks for documenting representativeness and intended use, while fairness and explainability audits operate at the model level rather than the dataset level. The four-dimension audit should be regarded as a minimal necessary check before stronger modeling claims are made, rather than as a comprehensive quality certification.

\section{Conclusion}

This study asked whether public educational prediction benchmarks are structurally reliable enough to support generalizable modeling claims. Across seven public datasets, the answer is usually no. OULAD, Dropout, and xAPI-Edu pass all four readiness checks, whereas the remaining four datasets fail at least one criterion and most fail the metadata-adequacy dimension. The main findings are:

(1) \textbf{Pervasive cross-group generalization failure.} Two datasets show structural fragility (Higher Ed: $R^2 < -8$ across all models; UCI Student: near-zero holdout), Entrance Exam shows linear-model-only failure ($-14\%$ retention, partly remediable by ensemble models), xAPI-Edu passes after correcting the feature-encoding artifact, and MM-TBA lacks grouping metadata entirely.

(2) \textbf{Structural collapse versus model-dependent failure.} On structurally fragile datasets, ensemble models amplify instability or collapse equally under group holdout; on Entrance Exam, they recover group-holdout performance (RF $R^2 \approx 0.32$ vs. linear $-0.060$).

(3) \textbf{Structured quality gradient.} The audit assigns profiles from Fragile to Strong, with each level corresponding to different remediation needs; stronger profiles cluster in larger, better-grouped releases with proximal features.

(4) \textbf{Random-split overstatement.} The iid-to-group collapse gap is small for Strong datasets but substantial for those with metadata-adequacy failures, with Higher Ed ($0.041$ to $-8.79$), UCI Student ($0.242$ to $-0.097$), and Entrance Exam ($0.438$ to $-0.060$) showing that random-split performance can overstate deployable signal.

(5) \textbf{Model-dependent group holdout.} Regularized and ensemble models sometimes survive group holdout where linear regression collapses, indicating that metadata adequacy should be evaluated jointly with model capacity.

(6) \textbf{Metric robustness.} A classification-metric sensitivity check confirms that audit profiles are robust to metric choice.

For reuse, the procedure is summarized as a four-step audit checklist:

(1) \textbf{Verify baseline gap}: does the proposed predictor beat a trivial baseline by a margin that is practically meaningful and reproducible across resampled splits?

(2) \textbf{Stress-test split instability}: does between-split variability remain smaller than the reported gain ($I < 1$)?

(3) \textbf{Check null separation}: are permutation-based significance tests consistent across admissible splits?

(4) \textbf{Audit metadata adequacy}: does performance survive group-aware holdout, and are the provenance fields needed for leakage-resistant validation present in the release?

The audit points to the following operational decision criteria:

\begin{table}[H]
\centering
\caption{Operational decision criteria derived from the seven-dataset audit. Thresholds are empirically motivated and intended as practical starting points.}
\label{tab:decision_criteria}
\resizebox{\textwidth}{!}{%
\begin{tabular}{p{2.5cm} p{2.8cm} p{2.5cm} p{4.5cm}}
\toprule
Audit dimension & Pass criterion & Fail indicator & Recommended action on failure \\
\midrule
Baseline gap & $\Delta_{\mathrm{MAE}} > 0$ with beat rate $\geq 0.90$ across $\geq 50$ splits & Beat rate $< 0.80$ or $\Delta_{\mathrm{MAE}}$ smaller than cross-split SD & Do not escalate to complex models; investigate target quality and feature relevance \\

Split instability & $I < 1.0$ (model variability smaller than gain) & $I \geq 1.0$ & Increase sample size, reduce feature dimensionality, or improve split design \\

Null separation & Permutation $p < 0.05$ consistent across $\geq 80\%$ of splits & Inconsistent or non-significant separation & Treat as insufficient evidence; do not publish benchmark claims \\

Metadata adequacy & Group holdout retains $\geq 50\%$ of iid $R^2$ under the linear reference model & $R^2$ collapse under group holdout & Add grouping metadata; run group-aware validation before deployment \\
\bottomrule
\end{tabular}}%
\end{table}

If all four checks pass, the dataset supports cautious benchmark claims. If one or two fail, targeted remediation may be sufficient. If three or more fail, the dataset is better treated as a research-stage resource requiring substantial revision before benchmark conclusions are drawn.

The contribution is not the checklist itself, but the finding that many public educational prediction benchmarks fail for a structural reason: their apparent signal does not survive stronger tests of cross-group generalization. The protocol exposes this reliability problem before benchmark claims are treated as evidence of generalizable educational AI performance.

\section*{Author contributions statement}
Conceptualization, Y.M. and L.Z.; methodology, Y.M. and L.Z.; software, L.Z.;
validation, Y.M. and L.Z.; formal analysis, Y.M.; investigation, Y.M.; data curation,
Y.M. and L.Z.; visualization, L.Z.; writing---original draft preparation, Y.M.;
writing---review and editing, Y.M. and L.Z.; supervision, L.Z. All authors reviewed
and approved the manuscript.

\section*{Funding}
This study was funded by the Education Department of Hunan Province, China.
Project Name: Empirical Study on the Impact of Changes in College English Classroom
Environment on Students' Engagement, Project Number: HNJG-2022-0608. The audit
methodology developed in this study supports the empirical evaluation of classroom
engagement prediction models funded under this grant.

\section*{Acknowledgements}
The authors thank the teams behind the seven audited datasets for their commitment
to open data.

\section*{Additional information}
Supplementary information accompanies this paper.

Ethics statement: According to the institutional policy governing secondary analysis
of existing public datasets at Hunan Agricultural University, ethical review and
approval were waived for this study because it used existing public dataset releases,
involved no direct interaction with human participants, and analyzed no newly
collected personal data.

Competing interests: The authors declare no competing interests.

\section*{Data availability statement}
All seven datasets are publicly available from their original
maintainers. The audit protocol implementation and analysis scripts are available at
\url{https://github.com/zhanglizhuo/BehaviorAudit} and archived with a permanent
DOI at Zenodo: \url{https://doi.org/10.5281/zenodo.21887892}.

Correspondence and requests for materials should be addressed to Lizhuo Zhang
(\url{zhanglizhuo@hunau.edu.cn}).

\section*{Abbreviations}
\noindent
\begin{tabular}{@{}ll}
MM-TBA & Multi-Modal Dataset for Teacher Behavior Analysis \\
OULAD  & Open University Learning Analytics Dataset \\
MAE    & Mean Absolute Error \\
EDM    & Educational Data Mining \\
RF     & Random Forest \\
GBT    & Gradient Boosting \\
\end{tabular}

\bibliography{behavioraudit}

\clearpage
\begingroup
\renewcommand{\thefigure}{S\arabic{figure}}
\renewcommand{\thetable}{S\arabic{table}}
\setcounter{figure}{0}
\setcounter{table}{0}
\renewcommand{\figurename}{Supplementary Figure}
\renewcommand{\tablename}{Supplementary Table}

\section*{Supplementary Materials}

This document contains supplementary tables and figures for the main manuscript.

\section*{S1 Model-Complexity Ablation}

\begin{table}[!h]
\centering
\caption{Model-complexity ablation: instability ratio ($I$) and beat rate across linear and nonlinear models on 100 repeated 80/20 splits.}
\label{tab:model_complexity}
\small
\begin{tabular}{l l c c c c c}
\toprule
Dataset & Model & MAE (mean $\pm$ SD) & $R^2$ (mean) & $r$ (mean) & $I$ & Beat rate \\
\midrule
MM-TBA & Linear & $0.795 \pm 0.089$ & $-0.067$ & $0.178$ & $2.21$ & $0.80$ \\
 & Ridge & $0.793 \pm 0.089$ & $-0.061$ & $0.183$ & $2.09$ & $0.81$ \\
 & Random Forest & $0.815 \pm 0.081$ & $-0.063$ & $0.196$ & $3.96$ & $0.65$ \\
 & Gradient Boosting & $0.856 \pm 0.095$ & $-0.227$ & $0.129$ & $4.60$ & $0.45$ \\
\midrule
Higher Ed & Linear & $1.695 \pm 0.199$ & $0.041$ & $0.469$ & $0.90$ & $0.83$ \\
 & Ridge & $1.686 \pm 0.195$ & $0.054$ & $0.470$ & $0.84$ & $0.83$ \\
 & Random Forest & $1.193 \pm 0.130$ & $0.521$ & $0.744$ & $0.18$ & $1.00$ \\
 & Gradient Boosting & $1.273 \pm 0.146$ & $0.464$ & $0.711$ & $0.23$ & $1.00$ \\
\midrule
xAPI-Edu & Linear & $0.363 \pm 0.025$ & $0.614$ & $0.786$ & $0.12$ & $1.00$ \\
 & Ridge & $0.357 \pm 0.023$ & $0.626$ & $0.799$ & $0.11$ & $1.00$ \\
 & Random Forest & $0.308 \pm 0.023$ & $0.682$ & $0.830$ & $0.09$ & $1.00$ \\
 & Gradient Boosting & $0.327 \pm 0.024$ & $0.663$ & $0.818$ & $0.10$ & $1.00$ \\
\midrule
Entrance Exam & Linear & $0.598 \pm 0.033$ & $0.438$ & $0.670$ & $0.12$ & $1.00$ \\
 & Ridge & $0.594 \pm 0.032$ & $0.448$ & $0.676$ & $0.11$ & $1.00$ \\
 & Random Forest & $0.612 \pm 0.036$ & $0.372$ & $0.628$ & $0.14$ & $1.00$ \\
 & Gradient Boosting & $0.595 \pm 0.033$ & $0.442$ & $0.671$ & $0.12$ & $1.00$ \\
\midrule
UCI Student & Linear & $2.011 \pm 0.148$ & $0.242$ & $0.530$ & $0.36$ & $1.00$ \\
 & Ridge & $2.010 \pm 0.148$ & $0.263$ & $0.530$ & $0.36$ & $1.00$ \\
 & Random Forest & $2.005 \pm 0.151$ & $0.285$ & $0.548$ & $0.37$ & $1.00$ \\
 & Gradient Boosting & $2.046 \pm 0.151$ & $0.257$ & $0.531$ & $0.41$ & $1.00$ \\
\midrule
Dropout & Linear & $0.206 \pm 0.006$ & $0.651$ & $0.807$ & $0.021$ & $1.00$ \\
 & Ridge & $0.206 \pm 0.006$ & $0.651$ & $0.807$ & $0.021$ & $1.00$ \\
 & Random Forest & $0.153 \pm 0.006$ & $0.684$ & $0.828$ & $0.019$ & $1.00$ \\
 & Gradient Boosting & $0.160 \pm 0.006$ & $0.692$ & $0.832$ & $0.019$ & $1.00$ \\
\midrule
OULAD & Linear & $0.296 \pm 0.002$ & $0.471$ & $0.687$ & $0.010$ & $1.00$ \\
 & Ridge & $0.296 \pm 0.002$ & $0.471$ & $0.687$ & $0.010$ & $1.00$ \\
 & Random Forest & $0.193 \pm 0.003$ & $0.577$ & $0.761$ & $0.009$ & $1.00$ \\
 & Gradient Boosting & $0.203 \pm 0.002$ & $0.610$ & $0.781$ & $0.008$ & $1.00$ \\
\bottomrule
\end{tabular}
\end{table}

\clearpage
\section*{S2 EDA Summary}

\begin{table}[!h]
\centering
\caption{Exploratory data analysis summary for the seven audited datasets. All releases contain zero missing values. The datasets that fail the audit (MM-TBA, Entrance Exam, Higher Ed, UCI Student) each fail at least one dimension, yet none show distinctive EDA red flags---such as extreme imbalance or redundant features---that would pre-emptively disqualify them; the fragility is only revealed through the four-dimension audit.}
\label{tab:eda_summary}
\small
\begin{tabular}{l r r l l r r r}
\toprule
Dataset & $N$ & Feat. & Target & Dist. & $n_{\mathrm{grp}}$ & Grp sz (min/mean/max) \\
\midrule
OULAD & 32,593 & 44 & Pass=1, Fail/W=0 & 0.47:0.53 & 22 & 365/1482/2498 \\
Dropout & 3,630 & 36 & Grad=1, Dropout=0 & 0.61:0.39 & 17 & 9/214/666 \\
Entrance Exam & 666 & 49 & 4-class ordinal & 24\%:32\%:30\%:15\% & 3 & 21/222/396 \\
UCI Student & 649 & 56 & Grade $G3$ [0,20] & $\mu=11.9$, $\sigma=3.2$ & 2 & 226/324/423 \\
xAPI-Edu & 480 & 72 & 3-class ordinal & 26\%:44\%:30\% & 12 & 19/40/95 \\
MM-TBA & 186 & 13 & Lecture quality [0.5,7.5] & $\mu=6.0$, $\sigma=1.0$ & 0 & --- \\
Higher Ed & 145 & 31 & Grade [0,7] & $\mu=3.5$, $\sigma=2.2$ & 9 & 2/16/66 \\
\bottomrule
\end{tabular}
\end{table}

\clearpage
\section*{S3 Train-Test $R^2$ Gap}

\begin{table}[!h]
\centering
\caption{Train-test $R^2$ gap across 100 repeated 80/20 splits. Within each split, Z-score parameters are fitted exclusively on the training partition and applied to the test partition. The gap (train $R^2$ -- test $R^2$) quantifies how much of the apparent iid signal is lost on held-out samples from the same distribution. On UCI Student, the linear-model train-test gap (0.105) is small relative to the iid-to-group-holdout collapse (0.242 to $-0.097$), confirming that the group-holdout failure is not attributable to ordinary overfitting. On OULAD, both gaps remain small.}
\label{tab:train_test_gap}
\small
\begin{tabular}{l l r r r r}
\toprule
Dataset & Model & Train $R^2$ mean & Test $R^2$ mean & Gap mean & Gap SD \\
\midrule
UCI Student & Linear & 0.347 & 0.242 & 0.105 & 0.091 \\
 & Ridge & 0.347 & 0.242 & 0.105 & 0.091 \\
 & RF & 0.895 & 0.265 & 0.630 & 0.084 \\
 & GBT & 0.603 & 0.259 & 0.345 & 0.090 \\
\midrule
Higher Ed & Linear & 0.460 & $-0.092$ & 0.552 & 0.335 \\
 & Ridge & 0.459 & $-0.079$ & 0.539 & 0.328 \\
 & RF & 0.877 & 0.085 & 0.793 & 0.143 \\
 & GBT & 0.850 & 0.054 & 0.796 & 0.171 \\
\midrule
OULAD & Linear & 0.471 & 0.464 & 0.008 & 0.026 \\
 & Ridge & 0.471 & 0.464 & 0.008 & 0.026 \\
 & RF & 0.933 & 0.573 & 0.360 & 0.024 \\
 & GBT & 0.629 & 0.587 & 0.042 & 0.026 \\
\bottomrule
\end{tabular}
\end{table}

\clearpage
\section*{S4 Group-Holdout Uncertainty}

\begin{table}[!h]
\centering
\caption{Across-group variation in group-holdout $R^2$. The SD quantifies how consistent the group-holdout performance is across held-out groups. For UCI Student (2 schools), the linear-model SD of 0.109 with a narrow range ($-0.206$ to $0.012$) confirms uniform collapse rather than group-specific failure. For OULAD (22 cohorts), the across-cohort SD of 0.162 and full range ($0.041$ to $0.654$) reflect positive---though variable---cross-group generalization.}
\label{tab:group_holdout_uncertainty}
\small
\begin{tabular}{l l r r r r}
\toprule
Dataset & Model & Groups tested & Mean $R^2$ & SD $R^2$ & Min / Max $R^2$ \\
\midrule
UCI Student & Linear & 2 & $-0.097$ & 0.109 & $-0.206$ / 0.012 \\
 & Ridge & 2 & $-0.094$ & 0.106 & $-0.201$ / 0.012 \\
 & RF & 2 & 0.018 & 0.109 & $-0.091$ / 0.127 \\
 & GBT & 2 & $-0.057$ & 0.136 & $-0.193$ / 0.079 \\
\midrule
Higher Ed & Linear & 4 & $-8.79$ & 6.39 & $-18.65$ / $-0.975$ \\
 & Ridge & 4 & $-8.58$ & 6.27 & $-18.35$ / $-0.969$ \\
 & RF & 4 & $-13.0$ & 14.0 & $-35.59$ / $-0.169$ \\
 & GBT & 4 & $-10.6$ & 11.3 & $-29.11$ / $-0.597$ \\
\midrule
OULAD & Linear & 22 & 0.410 & 0.162 & 0.041 / 0.654 \\
 & Ridge & 22 & 0.410 & 0.162 & 0.041 / 0.654 \\
 & RF & 22 & 0.482 & 0.211 & $-0.052$ / 0.747 \\
 & GBT & 22 & 0.533 & 0.165 & 0.039 / 0.752 \\
\bottomrule
\end{tabular}
\end{table}

\clearpage
\section*{S5 Linear-Model Coefficient Stability}

\begin{table}[!h]
\centering
\caption{Top-10 linear-model standardized coefficient stability across 100 repeated 80/20 splits for the three illustrative datasets. The sign-flip rate is the proportion of splits where the coefficient sign differs from the sign of the mean coefficient. On OULAD, all top-10 features show zero sign flips with low SD, indicating stable directional relationships. On UCI Student, sign-flip rates are near zero for most top features but f9 shows a 3\% rate and f2 a 4\% rate, indicating occasional directional instability; SDs are proportionally larger than on OULAD. On Higher Ed, sign-flip rates are near zero for most top features (f9: 2\%) with SDs reflecting greater uncertainty about effect magnitude.}
\label{tab:linear_coef_stability}
\small
\begin{tabular}{l l r r r}
\toprule
Dataset & Feature & Mean coef & SD coef & Sign-flip rate \\
\midrule
UCI Student & f5 & $-0.862$ & 0.063 & 0.00 \\
 & f4 & 0.393 & 0.057 & 0.00 \\
 & f49 & 0.270 & 0.032 & 0.00 \\
 & f48 & $-0.270$ & 0.032 & 0.00 \\
 & f11 & $-0.253$ & 0.051 & 0.00 \\
 & f30 & 0.229 & 0.061 & 0.00 \\
 & f33 & $-0.217$ & 0.054 & 0.00 \\
 & f9 & $-0.202$ & 0.097 & 0.03 \\
 & f2 & 0.199 & 0.086 & 0.04 \\
 & f22 & 0.190 & 0.047 & 0.00 \\
\midrule
Higher Ed & f28 & 0.669 & 0.123 & 0.00 \\
 & f0  & $-0.662$ & 0.134 & 0.00 \\
 & f17 & 0.618 & 0.121 & 0.00 \\
 & f1  & 0.506 & 0.119 & 0.00 \\
 & f8  & $-0.418$ & 0.103 & 0.00 \\
 & f25 & 0.402 & 0.119 & 0.00 \\
 & f4  & 0.401 & 0.132 & 0.00 \\
 & f2  & 0.332 & 0.108 & 0.00 \\
 & f15 & $-0.304$ & 0.103 & 0.00 \\
 & f9  & 0.300 & 0.109 & 0.02 \\
\midrule
OULAD & f3 & 0.175 & 0.005 & 0.00 \\
 & f2 & 0.162 & 0.006 & 0.00 \\
 & f4 & 0.114 & 0.003 & 0.00 \\
 & f1 & $-0.080$ & 0.005 & 0.00 \\
 & f0 & $-0.039$ & 0.003 & 0.00 \\
 & f5 & $-0.024$ & 0.003 & 0.00 \\
 & f10 & 0.022 & 0.001 & 0.00 \\
 & f11 & $-0.022$ & 0.001 & 0.00 \\
 & f29 & $-0.015$ & 0.002 & 0.00 \\
 & f27 & 0.013 & 0.002 & 0.00 \\
\bottomrule
\end{tabular}
\end{table}

\clearpage
\section*{S6 Random Forest Importance Stability}

\begin{table}[H]
\centering
\caption{Top-10 Random Forest feature importance stability across 100 repeated 80/20 splits. The coefficient of variation (CV = SD / mean) quantifies relative uncertainty. On OULAD, all top-10 CVs are below 0.12, indicating stable importance rankings. On UCI Student, CVs range from 0.08 to 0.20; on Higher Ed, from 0.18 to 0.41, reflecting greater split sensitivity in smaller datasets.}
\label{tab:rf_importance_stability}
\small
\begin{tabular}{l l r r r}
\toprule
Dataset & Feature & Mean importance & SD importance & CV \\
\midrule
UCI Student & f5 & 0.193 & 0.015 & 0.076 \\
 & f12 & 0.068 & 0.007 & 0.100 \\
 & f8 & 0.047 & 0.006 & 0.125 \\
 & f0 & 0.044 & 0.005 & 0.107 \\
 & f7 & 0.039 & 0.005 & 0.121 \\
 & f4 & 0.038 & 0.006 & 0.145 \\
 & f2 & 0.036 & 0.004 & 0.123 \\
 & f10 & 0.036 & 0.005 & 0.130 \\
 & f9 & 0.036 & 0.007 & 0.198 \\
 & f6 & 0.034 & 0.005 & 0.155 \\
\midrule
Higher Ed & f28 & 0.131 & 0.028 & 0.212 \\
 & f1  & 0.073 & 0.030 & 0.409 \\
 & f3  & 0.064 & 0.016 & 0.243 \\
 & f27 & 0.053 & 0.014 & 0.255 \\
 & f10 & 0.046 & 0.010 & 0.211 \\
 & f12 & 0.044 & 0.008 & 0.178 \\
 & f11 & 0.040 & 0.009 & 0.212 \\
 & f16 & 0.039 & 0.008 & 0.205 \\
 & f0  & 0.038 & 0.010 & 0.255 \\
 & f7  & 0.037 & 0.011 & 0.291 \\
\midrule
OULAD & f3 & 0.512 & 0.006 & 0.012 \\
 & f4 & 0.137 & 0.004 & 0.029 \\
 & f2 & 0.071 & 0.003 & 0.044 \\
 & f5 & 0.064 & 0.002 & 0.029 \\
 & f1 & 0.059 & 0.002 & 0.037 \\
 & f0 & 0.012 & 0.001 & 0.073 \\
 & f29 & 0.007 & 0.001 & 0.082 \\
 & f27 & 0.006 & 0.001 & 0.080 \\
 & f35 & 0.006 & 0.001 & 0.093 \\
 & f36 & 0.005 & 0.001 & 0.096 \\
\bottomrule
\end{tabular}
\end{table}

\clearpage
\section*{S7 Feature-Name Mapping}

\begin{table}[H]
\centering
\caption{Mapping from feature codes used in Supplementary Tables~S5--S6 to human-readable feature names for the three datasets included in the feature-attribution stability analysis. Feature codes correspond to zero-indexed column positions in the preprocessed feature matrix after one-hot encoding. Group-identifier columns (school for UCI Student, Course ID for Higher Ed) are excluded from the feature matrix, so their corresponding features do not appear in the tables.}
\label{tab:feature_mapping}
\small
\begin{tabular}{l l l}
\toprule
Dataset & Code & Feature name \\
\midrule
UCI Student & f0 & age \\
 & f2 & Fedu \\
 & f4 & studytime \\
 & f5 & failures \\
 & f6 & famrel \\
 & f7 & freetime \\
 & f8 & goout \\
 & f9 & Dalc \\
 & f10 & Walc \\
 & f11 & health \\
 & f12 & absences \\
 & f22 & Mjob\_health \\
 & f30 & Fjob\_teacher \\
 & f33 & reason\_other \\
 & f48 & higher\_no \\
 & f49 & higher\_yes \\
\midrule
Higher Ed & f0 & Student Age \\
 & f1 & Sex \\
 & f2 & Graduated high-school type \\
 & f3 & Scholarship type \\
 & f4 & Additional work \\
 & f7 & Total salary if available \\
 & f8 & Transportation to the university \\
 & f9 & Accomodation type in Cyprus \\
 & f10 & Mother's education \\
 & f11 & Father's education \\
 & f12 & Number of sisters/brothers (if available) \\
 & f15 & Father's occupation \\
 & f16 & Weekly study hours \\
 & f17 & Reading frequency (non-scientific books/journals) \\
 & f25 & Listening in classes \\
 & f27 & Flip-classroom \\
 & f28 & Cumulative grade point average in the last semester (/4.00) \\
\midrule
OULAD & f0 & num\_of\_prev\_attempts \\
 & f1 & vle\_total\_clicks \\
 & f2 & vle\_active\_weeks \\
 & f3 & assessment\_count \\
 & f4 & assessment\_score\_mean \\
 & f5 & assessment\_score\_std \\
 & f10 & gender\_F \\
 & f11 & gender\_M \\
 & f27 & highest\_education\_A Level or Equivalent \\
 & f29 & highest\_education\_Lower Than A Level \\
 & f35 & imd\_band\_20--30\% \\
 & f36 & imd\_band\_30--40\% \\
\bottomrule
\end{tabular}
\end{table}

\clearpage
\section*{S8 Cross-Model Attribution Agreement}

\begin{table}[H]
\centering
\caption{Cross-model Spearman rank correlation between linear-model absolute standardized coefficients and Random Forest impurity-based importances across 100 repeated 80/20 splits. $\rho_{\mathrm{all}}$ uses all features; $\rho_{\mathrm{top10}}$ uses the top-10 features by linear absolute coefficient in each split. On OULAD, cross-model agreement is high and stable across splits ($\rho_{\mathrm{top10}} = 0.817$, SD 0.065). On UCI Student and Higher Ed, agreement is weaker and more variable, consistent with their lower-readiness profiles.}
\label{tab:cross_model_spearman}
\small
\begin{tabular}{l r r r r r}
\toprule
Dataset & $p$ & \multicolumn{2}{c}{$\rho_{\mathrm{all}}$} & \multicolumn{2}{c}{$\rho_{\mathrm{top10}}$} \\
\cmidrule(lr){3-4} \cmidrule(lr){5-6}
 & & Mean & SD & Mean & SD \\
\midrule
UCI Student & 54 & 0.496 & 0.093 & 0.516 & 0.242 \\
Higher Ed & 30 & 0.285 & 0.126 & 0.498 & 0.245 \\
OULAD & 44 & 0.752 & 0.048 & 0.817 & 0.065 \\
\bottomrule
\end{tabular}
\end{table}

\clearpage
\section*{S9 Supplementary Figures}

\begin{figure}[!h]
\centering
\includegraphics[width=0.95\textwidth]{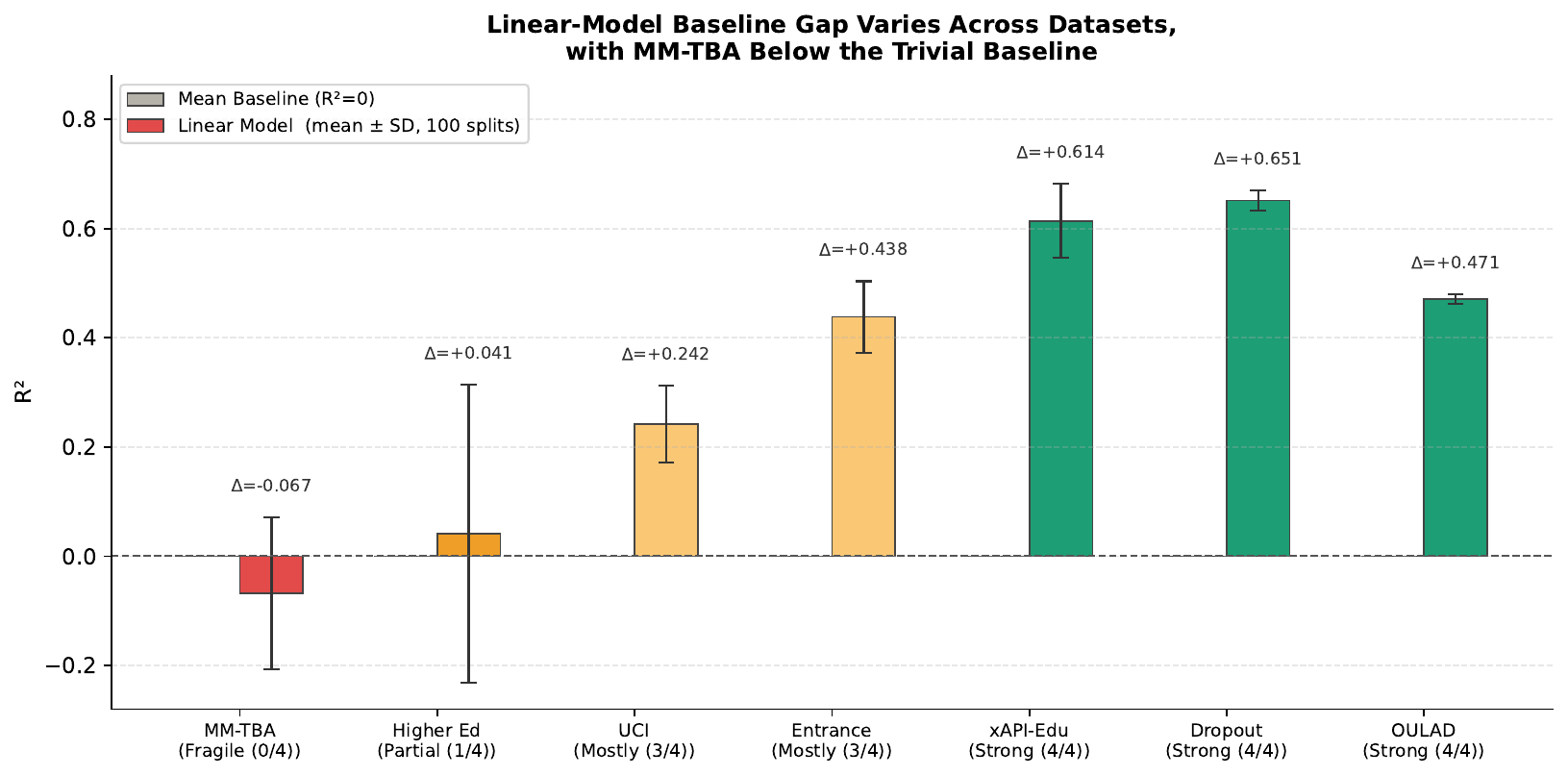}
\caption{Baseline gap: linear-model mean $R^2$ (with cross-split SD error bars) over 100 random 80/20 splits for the seven datasets, ordered by readiness profile. The grey reference bar denotes the trivial mean baseline ($R^2 = 0$). The figure complements Table~2 by showing the dispersion of split-level $R^2$ around the reported mean; MM-TBA and Higher~Ed have means near or below zero with relatively wide spread, consistent with their Fragile / Partial classifications.}
\label{fig:s_baseline_gap}
\end{figure}

\begin{figure}[!h]
\centering
\includegraphics[width=0.95\textwidth]{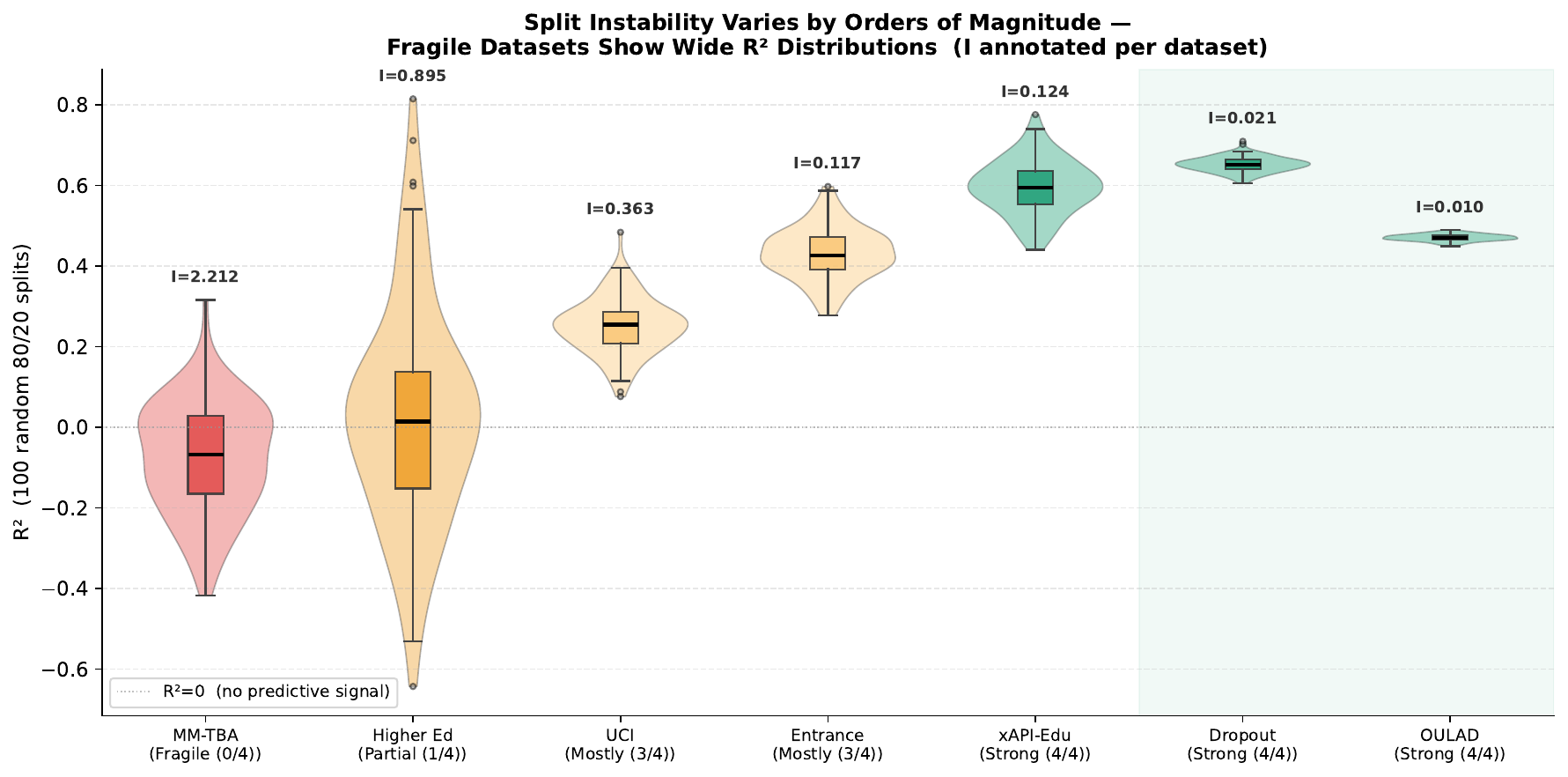}
\caption{Split instability: violin and box plots of the linear-model $R^2$ across 100 random 80/20 splits for each dataset, ordered by readiness profile. The instability ratio $I$ is annotated above each violin. Strong datasets (OULAD, Dropout) have tight distributions ($I < 0.03$); MM-TBA's distribution is broad and centred on negative $R^2$ ($I = 2.21$), reproducing Table~2 visually.}
\label{fig:s_split_instability}
\end{figure}

\clearpage

\begin{figure}[!h]
\centering
\includegraphics[width=0.95\textwidth]{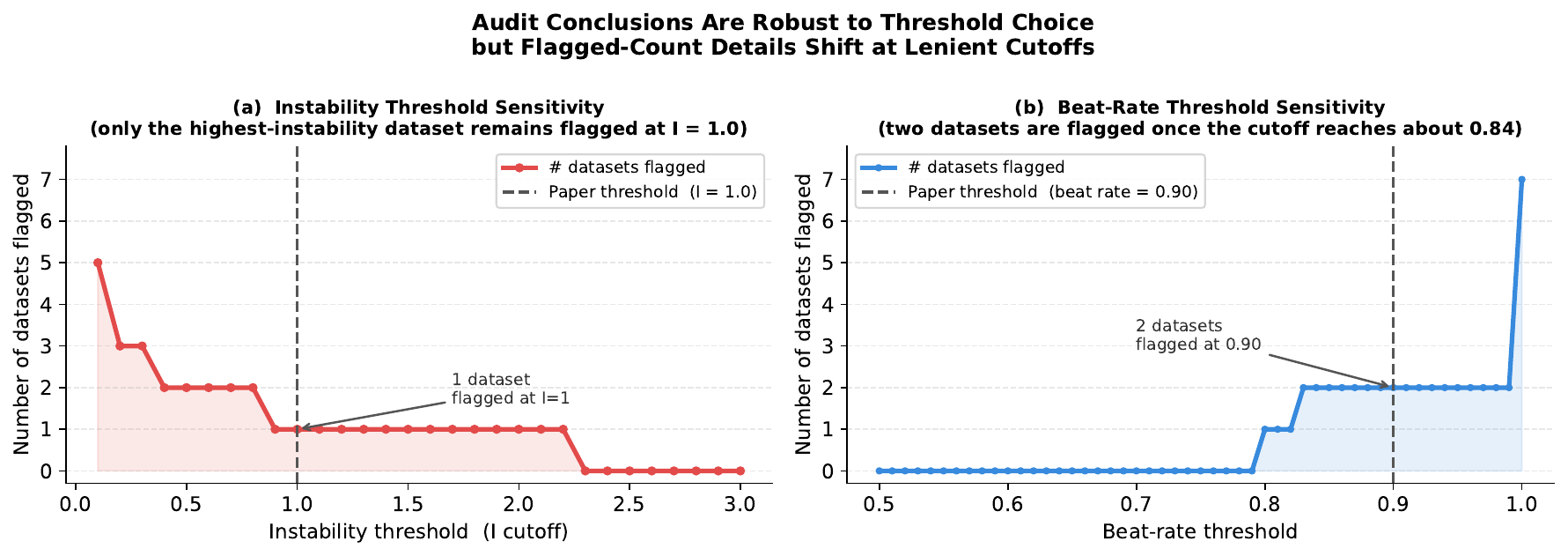}
\caption{Threshold-sensitivity sweeps for the two operational cutoffs used in the audit. Panel~(a) varies the instability cutoff over $I \in [0.1, 3.0]$ and reports the number of datasets that would be flagged as failing split instability; the paper's choice of $I = 1.0$ flags one dataset (MM-TBA, with linear $I = 2.21$), while more stringent cutoffs below about $0.9$ additionally flag Higher~Ed. Panel~(b) varies the beat-rate cutoff over $[0.5, 1.0]$; the paper's choice of $0.90$ flags two datasets, whereas more lenient cutoffs below about $0.84$ reduce the flagged count. The main qualitative pattern remains unchanged: MM-TBA is always the weakest case, and Higher~Ed is the only borderline dataset near the operational thresholds.}
\label{fig:s_threshold_sensitivity}
\end{figure}
\endgroup

\end{document}